\documentclass[lettersize,journal]{IEEEtran}
\usepackage{amsmath,amsfonts}
\usepackage{algorithmic}
\usepackage{algorithm}
\usepackage{array}
\usepackage[caption=false,font=normalsize,labelfont=sf,textfont=sf]{subfig}
\usepackage{textcomp}
\usepackage{stfloats}
\usepackage{url}
\usepackage{verbatim}
\usepackage{graphicx}
\usepackage{cite}

\usepackage[table]{xcolor}  % 在导言区添加
\usepackage{graphicx}
\usepackage{caption} % 引入 caption 宏包
\usepackage{multirow}
\usepackage{booktabs}

\begin{document}

\title{MoonGS: High-quality Representation of the Lunar Surface via Gaussian Splatting Using Robust Depth Features from Image Pairs}

% % \thanks{This paper was produced by the IEEE Publication Technology Group. They are in Piscataway, NJ.}% <-this % stops a space
% % \thanks{Manuscript received April 19, 2021; revised August 16, 2021.}
% }

\author{Yun Jiang,
        Bo Zheng,
        Yingying Zhang,
        Xueming Xiao,
        Tao Hu,
        Hutao Cui,
        Zhiguo Meng,
        Ke Gao,
        Yang Gao,
        and Meibao Yao
\thanks{Corresponding author: Meibao Yao.}%
\thanks{Yun Jiang, Meibao Yao, Zhiguo Meng, and Ke Gao are with Jilin University, Changchun, China.}%
\thanks{Yingying Zhang is with Beijing Institute of Control Engineering, Beijing, China.}
\thanks{Xueming Xiao is with Changchun University of Science and Technology, Changchun, China.}%
\thanks{Tao Hu and Bo Zheng are with Shanghai Aerospace Control Technology Institute, Shanghai, China.}%
\thanks{Hutao Cui is with Harbin Institute of Technology, Heilongjiang, China.}%
\thanks{Yang Gao is with The Hong Kong University of Science and Technology, Hong Kong, China.}%
}



\maketitle

\begin{abstract}
High-quality 3D reconstruction of lunar terrain from sparse rover images is indispensable for autonomous lunar exploration, but remains challenging because viewpoint overlap is insufficient, surface textures are weak, and data volume is limited. We propose MoonGS, the first feed-forward 3D Gaussian Splatting framework tailored to lunar scenes. Given only two input images, MoonGS predicts pixel-aligned Gaussian primitives in a single forward pass and renders photorealistic novel views without any per-scene optimization. 
MoonGS (i) adopts an adaptable backbone design that seamlessly integrates advanced vision foundation models to extract robust depth features;
(ii) integrates semantic priors in two manners: merging semantic cues with visual features to refine Gaussian parameter estimation, and adopting a semantic ranking loss that regularizes background depth; (iii) employs an entropy-guided heuristic resampling strategy to augment sparse observations by selecting the most informative distant viewpoints with negligible overhead. Experiments on the LuSNAR benchmark and our synthetic weak-texture MoonBlenderdataset show that MoonGS surpasses state-of-the-art feed-forward NeRF/3DGS baselines by \mbox{+4.9 dB}~PSNR, \mbox{+0.29}~SSIM, and \mbox{$\downarrow$40\%}~LPIPS while maintaining sub-second inference. 
Furthermore, we validate the broad applicability of our framework by demonstrating that it effectively leverages state-of-the-art backbones, including VGGT, to significantly boost performance.
Qualitative evaluations on Chang’e mission imagery also show best visual quality among compared methods, indicating robustness on real lunar data. The source code and dataset are publicly available at \url{https://github.com/InRobots/MoonBlender}.
\end{abstract}


\section{INTRODUCTION}

\begin{figure}[htb]
    \centering
    \begin{minipage}[t]{0.323\linewidth}
        \centering
        \caption*{Ref.1}
        \includegraphics[width=\linewidth]{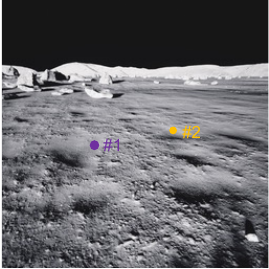}
    \end{minipage}
    \hfill
    \begin{minipage}[t]{0.323\linewidth}
        \centering
        \caption*{Ref.2}
        \includegraphics[width=\linewidth]{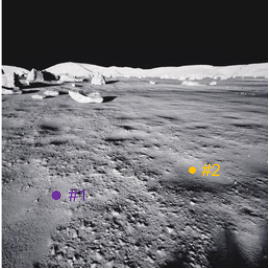}
        
    \end{minipage}
    \hfill
    \begin{minipage}[t]{0.323\linewidth}
        \centering
        \caption*{Ground Truth}
        \includegraphics[width=\linewidth]{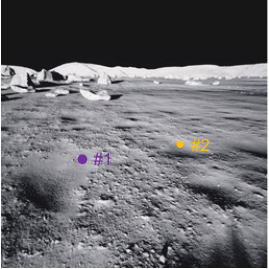}
        
    \end{minipage}
    \hfill
    % \vspace{1em}  % 两行之间的间距
    % 每张图分配一个 minipage
    \begin{minipage}[t]{0.323\linewidth}
        \centering
        \includegraphics[width=\linewidth]{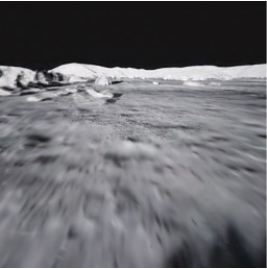}
        \caption*{pixelSplat\cite{charatan2024pixelsplat}}
    \end{minipage}
    \hfill
    \begin{minipage}[t]{0.323\linewidth}
        \centering
        \includegraphics[width=\linewidth]{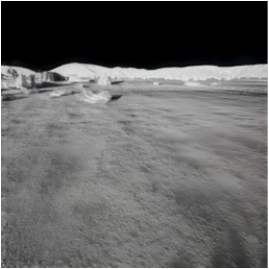}
        \caption*{MVSplat\cite{chen2024mvsplat}}
    \end{minipage}
    \hfill
    \begin{minipage}[t]{0.323\linewidth}
        \centering
        \includegraphics[width=\linewidth]{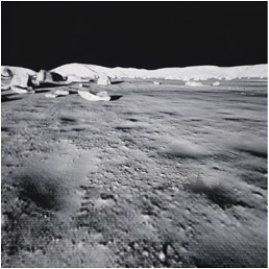}
        \caption*{Ours}
    \end{minipage}
    \vspace{-0.5em}
   \caption{\textbf{Novel view synthesis comparison.} Compared to existing 3DGS-based methods, our approach provides higher-fidelity novel view synthesis results.}
    \label{overall}
\end{figure}

Lunar exploration through robotic rovers is a cornerstone for future long term lunar residency activities. We investigate the problem of high-fidelity novel view synthesis from sparse image observations in lunar environments with higher real-time performance, as shown in Fig.~\ref{overall}, which can enable rovers to achieve comprehensive terrain understanding~\cite{4399223}, precise visual localization~\cite{mur2017orb}, and improved real-time decision-making. As an integral part of 3D scene reconstruction, novel view synthesis is a task dedicated to inferring previously unseen viewpoints of a scene from partial observations. Classical dense 3D reconstruction methods, including point cloud or voxel representations~\cite{vineet2015incremental, whelan2015elasticfusion, dai2017bundlefusion}, typically depend on local image similarity metrics. However, these approaches can hardly handle scenes containing repetitive textures or limited overlapping regions, resulting in poor-quality novel view synthesis. Recent advances in neural rendering, such as Neural Radiance Fields (NeRFs)~\cite{mildenhall2021nerf}, have the ability to model complex scenes and render realistic images. 3D Gaussian Splatting (3DGS)~\cite{kerbl20233d}, an emerging method extending NeRF, significantly accelerates rendering speed and improves efficiency, while still maintaining high fidelity. 
However, since 3DGS requires training from scratch for each new scene, it currently cannot meet real-time application requirements. Additionally, it requires densely captured images to ensure quality, making it unsuitable for common scenarios with extremely sparse observations in lunar exploration missions.

Feed-forward radiance field methods partially address the challenges posed by sparse data and limited real-time capability. For instance, NeRF-based approaches (e.g., MuRF~\cite{xu2024murf} and PixelNeRF~\cite{yu2021pixelnerf}) address these challenges by performing efficient inference directly from sparse observations. In comparison, feed-forward 3DGS methods enable more photorealistic and efficient novel view synthesis by explicitly utilizing 3D Gaussian scene representations. Leading scalable reconstruction models in this category (e.g., pixelSplat~\cite{charatan2024pixelsplat} and MVSplat~\cite{chen2024mvsplat}) employ feed-forward architectures to efficiently generate Gaussian primitives for scene representation. Despite recent progress, lunar scene reconstruction still faces several distinct and unresolved challenges: (i) limited viewpoint overlap during exploration often invalidates traditional epipolar constraints, hindering accurate cross-view feature matching; (ii) lunar background regions exhibit weaker textures compared to foreground objects, complicating depth estimation and geometric reconstruction due to sparse visual cues; (iii) mission-imposed resource constraints severely limit the number and quality of observations, further challenging the accurate representation of lunar surfaces.

\begin{figure}[t]
    \centering
    \includegraphics[width=\linewidth]{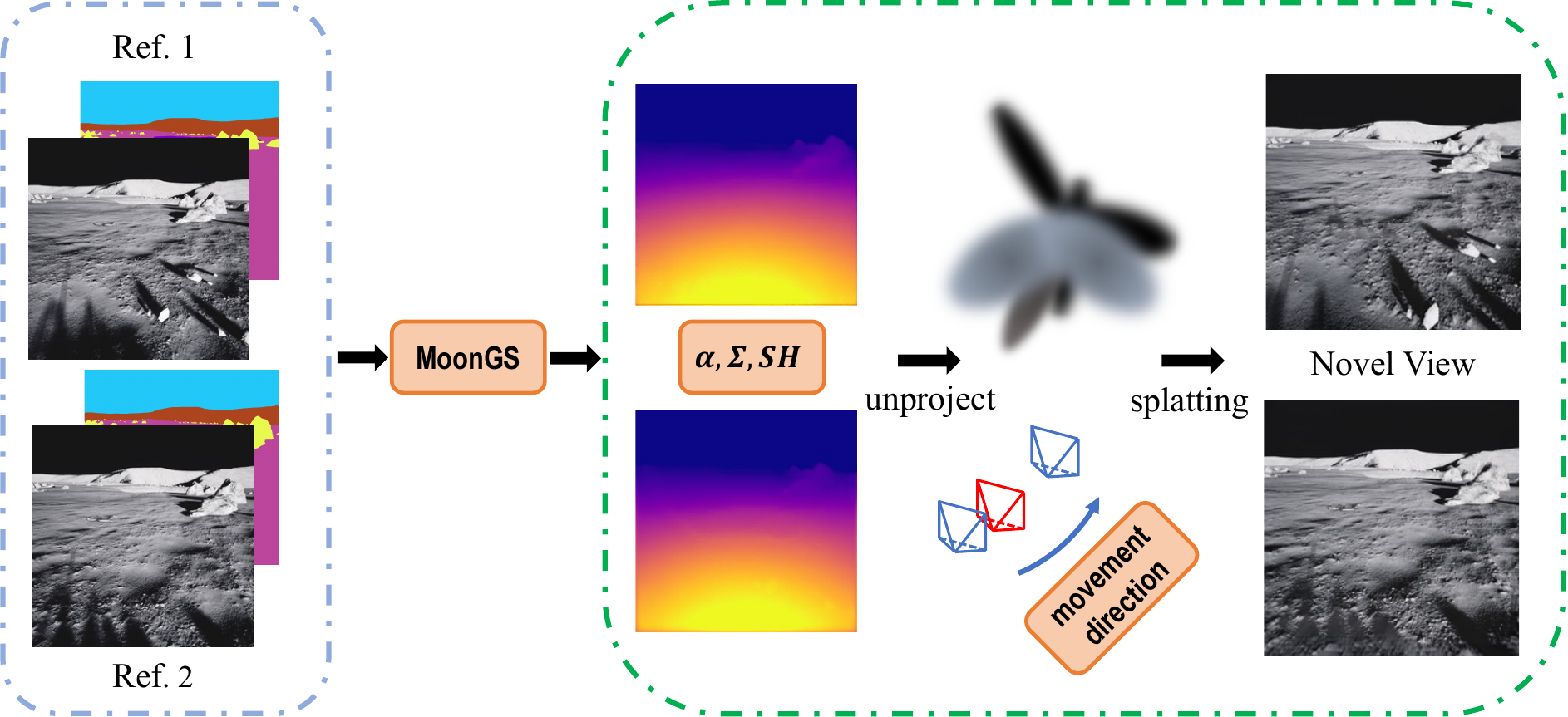}
    \caption{\textbf{Overview of MoonGS.} From a pair of input images and their associated semantic information, MoonGS performs forward inference to obtain the parameters of Gaussian primitives that represent the lunar surface scene. It then utilizes a camera with a known pose to unproject these Gaussian primitives onto the reliable depth positions. With this explicit representation, images can be rendered from camera viewpoints that have not been directly observed.}
    \label{fig:overview}
\end{figure}

To overcome these critical challenges, we propose MoonGS, a feed-forward 3DGS framework that synthesizes novel views from sparse lunar surface images and rich semantic information with high fidelity, while achieving real-time performance, as illustrated in Fig.~\ref{fig:overview}. To begin with, (i) MoonGS leverages robust cross-view depth features provided by an off-the-shelf vision foundation model to achieve a reliable scene representation, particularly in regions lacking sufficient viewpoint overlap. Moreover, (ii) to further enhance performance, we integrate semantic information into our method in two critical ways: firstly, semantic priors are used to improve the prediction of Gaussian parameters, resulting in more stable scene geometry; secondly, a semantic ranking loss is introduced to regularize depth estimation in weak-textured background regions (e.g., the sky), significantly reducing geometric errors. In addition, (iii) we propose a heuristic resampling strategy to effectively augment sparse lunar surface observations, substantially enhancing the fidelity of synthesized novel views. Meanwhile, (iv) we also introduce a new synthetic lunar dataset as an extension to existing lunar datasets, deliberately designed with extremely sparse sky textures to specifically evaluate reconstruction performance under weakly textured conditions.

In summary, our main contributions are as follows:
\begin{enumerate}
  \item We introduce MoonGS, a feed-forward 3DGS framework specifically designed for the Moon, which represents scenes using Gaussian primitives and enables high-quality synthesis of novel views from sparse viewpoints in real time.
  \item We fully exploit robust depth features from an off-the-shelf vision foundation model to address insufficient viewpoint overlap, effectively integrate semantic priors and a semantic ranking loss to handle unreliable depth estimation in weakly textured regions, and adopt a heuristic resampling strategy to overcome data sparsity.
  \item We provide a dataset named MoonBlender as an extension of existing lunar datasets, specifically designed to evaluate the performance of scene representation methods under weakly textured conditions.
\end{enumerate}

\section{Related Work}
\begin{figure*}[ht]
\centering
\includegraphics[width=\textwidth]{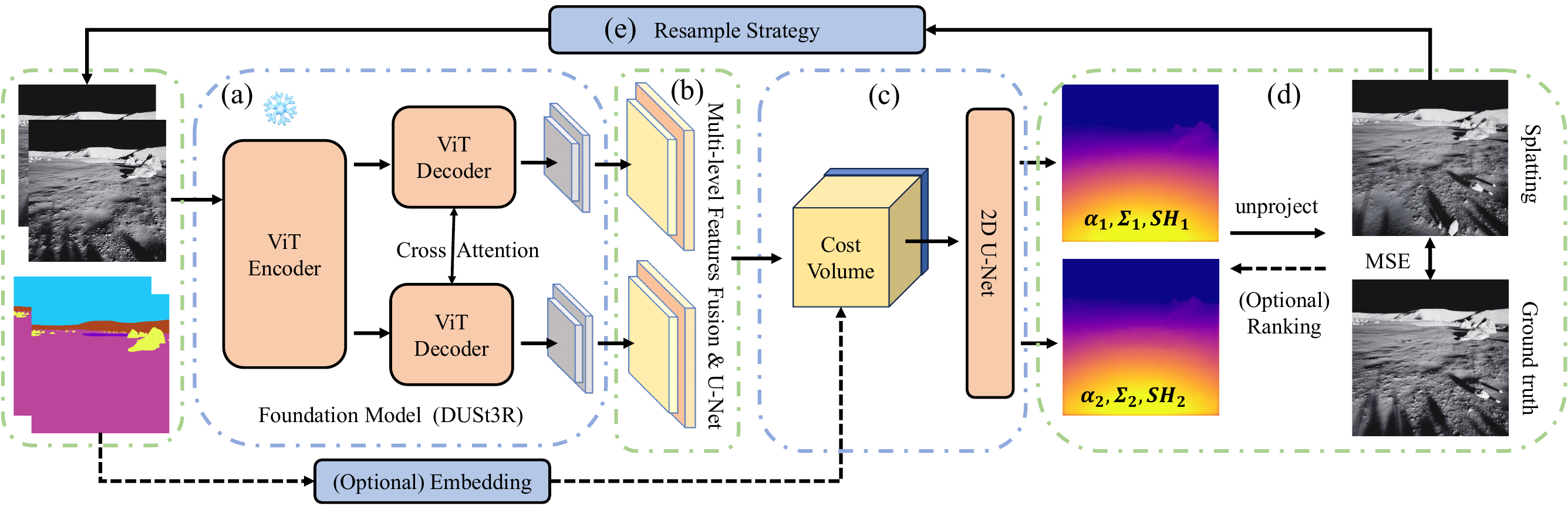}
\caption{\textbf{Architecture diagram.} (a) MoonGS utilizes DUSt3R to extract multi-level robust depth features. (b) Multi-level features are fused and refined through 2D U-Net. (c) The model constructs a cost volume using the extracted features and optionally integrates semantic embeddings as inputs for prediction to obtain depth maps and Gaussian parameters. (d) It then unprojects the Gaussian parameters into space, synthesizes novel views using Gaussian splatting, and employs MSE for supervised learning, with an optional ranking loss. (e) Finally, the model implements a resampling strategy to select additional positional images that can improve rendering quality as inputs. Dashed arrows denote optional components.}
\label{fig:architecture}
\vspace{-1em}
\end{figure*}

\subsection{Celestial Surface Reconstruction Based on Radiance Fields}
Reconstructing celestial surface environments is critical for tasks such as geological analysis~\cite{kang2018coarse} and navigation simulation~\cite{8701525}, providing essential data for robotic perception~\cite{10746318, 10610211,XIAO20192535, YANG20241327, 10210278, 9646256, 10360155, LIU2023492} and mission planning~\cite{XIONG2024167, dynamic_SLAM}. Early approaches to novel view synthesis were based on light fields, initially using densely sampled data~\cite{10.1145/3596711.3596759, 10.1145/383259.383309} and subsequently allowing unstructured capture~\cite{10.1145/3596711.3596764}. The advent of structure-from-motion (SfM)~\cite{10.1145/3596711.3596766} opened the door to synthesizing novel views from photo collections, by estimating a sparse point cloud during camera calibration. Building on SfM, multi-view stereo (MVS) techniques~\cite{Goesele2007MultiViewSF} enabled the development of full 3D reconstruction pipelines and further advanced view synthesis algorithms~\cite{10.1145/3272127.3275084, kopanas2021pointbasedneuralrenderingperview}. These methods typically reproject and blend input images based on the recovered geometry, producing high-quality results in many scenarios. However, they can still suffer from unreconstructed regions or artifacts caused by inaccuracies in the MVS-generated geometry. Neural radiance field methods achieve higher fidelity and have shown promise in novel view synthesis, 3D reconstruction, and pose estimation~\cite{10680902}. Mildenhall et al.\cite{mildenhall2021nerf} encode a scene as an MLP-based implicit function and use volume rendering, which is computationally expensive due to extensive ray sampling. Subsequent techniques\cite{muller2022instant, yu2021plenoctrees} accelerate radiance field methods, and among these, 3DGS~\cite{kerbl20233d} employs Gaussian primitives with a rasterization approach to circumvent ray sampling, reconstructing scenes from dense posed images.
Radiance fields have also been applied to celestial environments for high-fidelity visualization. For instance, MaRF~\cite{giusti2022marf} reconstructs Martian landscapes from NASA’s PDS, and Zhang et al.~\cite{zhang2024neural} extends neural radiance fields to lunar surface scenes. However, existing approaches often suffer from data sparsity and lack real-time performance. To address these issues in lunar applications, we format sparse lunar data as image pairs and design a specialized model architecture to exploit this sparsity. As a result, our method synthesizes high-quality lunar images in real time, strengthening robotic perception in lunar exploration.

\subsection{Feed-forward Radiance Fields}
To enable real-time reconstruction, researchers have explored feed-forward radiance field methods that reconstruct a scene from a single forward pass. PixelNeRF~\cite{yu2021pixelnerf} uses one or few input images to form an entire scene, integrating pixel features into rendering and showing that radiance fields can be learned from sparse views. Du et al.\cite{du2023learning} adopt epipolar sampling and a lighter renderer for better results, and MuRF\cite{xu2024murf} constructs a cone volume of the target view, achieving strong performance via convolutional neural networks. However, all these NeRF-based methods rely on ray sampling, resulting in slower performance.

3D Gaussian Splatting (3DGS)\cite{kerbl20233d} is significantly faster in both training and rendering. Consequently, many sparse-view reconstruction methods have adopted 3DGS. For instance, Splatter Image\cite{szymanowicz2024splatter} employs U-Net to directly infer pixel-wise Gaussian parameters from a single view for object-level reconstruction, while PixelSplat~\cite{charatan2024pixelsplat} learns epipolar attention from two related views to predict depth distributions. However, ambiguous feature-to-depth mappings can misposition Gaussians and blur the final images. MVSplat~\cite{chen2024mvsplat} constructs a cost volume for more precise geometry, and MVSGaussian~\cite{liu2024mvsgaussian} similarly uses cost volumes for depth estimation. Yet, end-to-end MVS approaches struggle with limited overlap in planetary data and are not tailored to these scenarios.

In contrast, MoonGS leverages a vision foundation model for depth-related features, inferring reliable depth even with sparse observations. It also uses rich lunar semantics as prior knowledge and employs a heuristic resampling strategy. These techniques yield superior reconstruction results for lunar scenes.

\subsection{Key Components of Feed-forward 3DGS}
In feed-forward 3DGS models, depth estimation is the most critical component, as it determines the positions of the Gaussians, which are more difficult to optimize compared with other parameters. Certain carefully designed approaches help obtain reliable depth estimates. Notably, pixelSplat~\cite{charatan2024pixelsplat} employs a combination of DINOv2~\cite{oquab2023dinov2} and an epipolar transformer for feature extraction, whereas MVSplat~\cite{chen2024mvsplat} uses UniMatch’s\cite{xu2023unifying} CNN and transformer modules to derive features. These components share the characteristic of first using the same encoder to acquire image encodings, followed by modules with cross-attention to decode and obtain cross-view features. However, DINOv2 is independent of depth estimation. UniMatch relies on matching between two input images to estimate depth, making it less robust in face of occlusions or other challenging scenarios. In contrast, vision foundation models such as DUSt3R~\cite{wang2024dust3r} and VGGT~\cite{wang2025vggtvisualgeometrygrounded} are pretrained on large-scale datasets and improve overall depth consistency and accuracy by generating pointmaps for each view. The features extracted from the foundational model further enhance performance in depth-related downstream tasks, such as image matching~\cite{leroy2024grounding}.
\section{Methods}
We propose MoonGS, a Gaussian-based sparse view scene reconstruction method, whose architecture is shown in Fig.\ref{fig:architecture}. Given a pair of images along with their corresponding camera parameters and semantic images, our approach infers the Gaussian parameters representing the entire scene using a feed-forward procedure. With this scene representation, we can render images from viewpoints that are not observed directly. The key to our approach lies in how we infer more reliable depth, enabling precise determination of the Gaussian primitives’ positions, even in non-overlapping areas. Subsection~\ref{p3dgs} clarifies our motivations and provides essential preliminaries on 3DGS~\cite{kerbl20233d}. Subsection~\ref{problem_def} introduces the problem definition. Subsection~\ref{robust_depth_features} explains how we use a vision foundation model to extract robust depth features, which is crucial for reconstructing a scene from sparse images. Subsection~\ref{semantic_methods} describes how we incorporate rich lunar semantic information to enhance the model's performance. Subsection~\ref{resample_strategy} designs a heuristic resampling strategy that can effectively guide the resampling positions of lunar surface operational equipment, thereby improving the quality of scene reconstruction.

\subsection{Motivations and Preliminaries}\label{p3dgs}
\begin{figure}[t]
    \centering
    \includegraphics[width=0.8\linewidth]{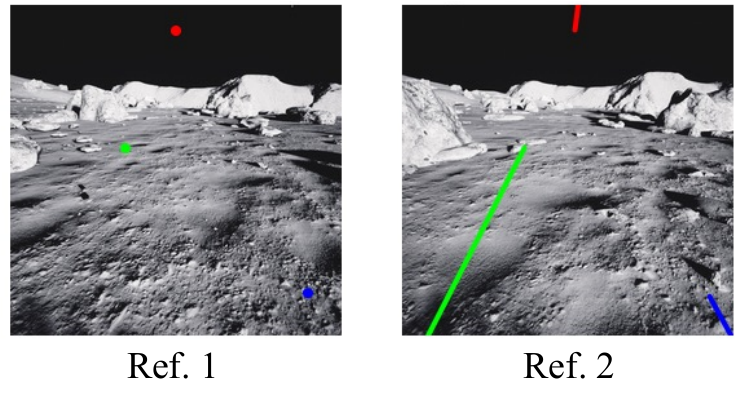}
    \vspace{-5pt}
    \caption{\textbf{Epipolar‐line visualization of three reconstruction challenges.} Three types of epipolar lines illustrate difficulties encountered during reconstruction in lunar environments. Ref.1 and Ref.2 depict two viewpoints of the same lunar scene. Colored points in Ref.1 correspond to regions indicated by matching-colored epipolar lines in Ref.2. The \textcolor{blue}{\textbf{blue}} lines denote regions with insufficient viewpoint overlap, resulting in invalid epipolar constraints. The \textcolor{red}{\textbf{red}} lines indicate weakly textured background regions, where epipolar constraints are prone to errors. The \textcolor{green}{\textbf{green}} lines mark areas with limited observations at a distance, making high-fidelity texture prediction challenging.}
    \vspace{-15pt}
    \label{fig:epline}
\end{figure}

\noindent \textbf{Motivations.} As lunar exploration missions progress, integrating 3DGS for scene reconstruction plays a crucial role in augmenting robotic perception. Due to the inherently sparse nature of lunar data, applying standard methods often leads to artifacts or even reconstruction failures. 
% Therefore, in situations where only sparse data are available, it is essential to focus on how to use pairs of images as input for reconstructing lunar scenes. 
Although several 3DGS-based methods exist for sparse-view reconstruction, they still face various difficulties. As illustrated in Fig.\ref{fig:epline}, data collected by a forward-moving lunar rover encounter multiple issues: invalid epipolar constraints in non-overlapping regions, insufficient observations for regions far from the camera, and ambiguous depth in low texture sky areas, among others. Although state-of-the-art approaches—such as pixelSplat~\cite{charatan2024pixelsplat} and MVSplat~\cite{chen2024mvsplat}—mitigate these challenges only partially, key problems remain unsolved. In this work, we aim to address these limitations more effectively.

% Advanced methods, such as pixelSplat~\cite{charatan2024pixelsplat} and MVSplat~\cite{chen2024mvsplat}, also encounter these problems. In this work, our aim is to address these issues effectively.

\noindent \textbf{Preliminaries.} 3DGS is a novel explicit scene representation method that employs anisotropic Gaussian primitives to model entire scenes. Each Gaussian primitive $G$ is defined by a position $\mu \in \mathbb{R}^3$, a covariance matrix $\Sigma$, a color $c \in \mathbb{R}^3$, and an opacity $\alpha \in \mathbb{R}$. The covariance matrix is parameterized by a scaling vector $S \in \mathbb{R}^3$ and a rotation matrix $R \in \mathbb{R}^{3 \times 3}$ as follows:
\begin{equation}
\Sigma = R \, S \, S^T \, R^T,
\label{eq:Sigma}
\end{equation}
where $S$ and $R$, respectively control the scale and orientation of the Gaussian.

Conceptually, the value of the Gaussian primitive at each position $x$ can be expressed as:
\begin{equation}
g(x) = \exp\!\Bigl(-\tfrac{1}{2}\,(x - \mu)^T \,\Sigma^{-1} \,(x - \mu)\Bigr).
\label{eq:Gx}
\end{equation}
3DGS~\cite{kerbl20233d} provides a highly efficient differentiable renderer, where the final color is computed through volumetric accumulation of opacity $\alpha$ as follows:
\begin{equation}
Color = \sum_{i \in N} c_i \,\alpha_i \, g(x_i) \,\prod_{j=1}^{i-1} \Bigl(1 - \alpha_j \, g(x_j)\Bigr),
\label{eq:C}
\end{equation}
where \( N \) denotes the number of accumulated Gaussian primitives. Unlike NeRFs~\cite{mildenhall2021nerf, muller2022instant}, which requires sampling along each ray, 3DGS supports efficient tile-based rendering, allowing the entire process to be parallelized on the GPU. In other words, our goal is to reconstruct an entire scene in a Gaussian representation, which entails estimating the full set of parameters for each Gaussian primitive.

\subsection{Problem Definition}\label{problem_def}

Given a pair of images \( \mathcal{I} = \{I_1, I_2\} \in  \mathbb{R}^{H \times W \times 3}\) and their associated camera parameters \( \mathcal{P} = \{P_1, P_2\} \), where \( P_i = K_i[R_i, t_i] \), \( K \) represents the camera intrinsic matrix, \( R \) represents the rotation matrix, and \( t \) represents the translation vector. Our goal is to design a learnable model, with input consisting of image pairs, camera parameters, and other available information, and outputs representing the Gaussian parameters of the entire scene. This process can be formulated as:
% \begin{equation}
% f_{\theta_{\text{train}}, \theta_{\text{fixed}}}(\{I_i, P_i, C_i\}_{i = 1,2}) = \{(\mu_j, \Sigma_j,c_j, \alpha_j) \}_{j=1}^{H \times W \times 2},
% \end{equation}
% \begin{equation}
% f_{\theta_{\text{train}}, \theta_{\text{fixed}}}\left(\{I_i, P_i, C_i\}_{i = 1,2}\right) = 
% \left\{
% \left(\mu_j, \Sigma_j, c_j, \alpha_j\right)
% \right\}_{j=1}^{H \times W \times 2},
% \end{equation}
\begin{equation}
\small
f_{\theta_{\text{train}}, \theta_{\text{fixed}}}\left(\{I_i, P_i, C_i\}_{i=1,2}\right) = \{(\mu_j, \Sigma_j, c_j, \alpha_j)\}_{j=1}^{H \times W \times 2},
\end{equation}
where \( f_{\theta} \) represents a feed-forward model. \( \theta_{\text{train}} \) denotes the learnable parameters in the model and \( \theta_{\text{fixed}} \) indicates the frozen parameters. \( C \) represents the prior information input to the model, such as semantic information. This model can ultimately obtain pixel-aligned Gaussian parameters, including position \( \mu \), opacity \( \alpha \), covariance \( \Sigma \), and color \( c \). Once the scene representation parameters are obtained, rendering from any viewpoint can be achieved using splatting~\cite{kerbl20233d}. The most critical factor in this process is the prediction of the Gaussian position, which is essential for effective training and high-quality rendering. The position parameters are typically obtained by predicting depth and unprojecting it using camera parameters. Therefore, the key issue of position prediction translates to the accuracy of depth prediction. Our method's solution lies in leveraging the vision foundation model to enhance depth predictability in lunar scenes. Precisely, the vision foundation model can provide robust depth features that can be used for better depth appraisal and form the basis for estimating other Gaussian parameters, establishing robust geometric priors even in non-overlapping regions.

\subsection{Robust Depth Features and Cost Volume}\label{robust_depth_features}
Our method focuses on how to obtain effective depth features to precisely locate Gaussian primitives. As a vision foundation model, DUSt3R~\cite{wang2024dust3r} is designed to extract cross-view features from a pair of input images. These features are subsequently decoded to generate a depth-related pointmap. Since DUSt3R has been pre-trained on a large amount of data, the features it extracts perform well in downstream tasks. Therefore, we treat the features extracted from DUSt3R as robust depth features, which constitute a core component of our approach. 

\noindent \textbf{Robust depth features extraction.} As shown in Fig.\ref{fig:architecture}, the vision foundation model (specifically DUSt3R) consists of two identical branches with the same structure, including an image encoder and decoder. Each branch only inputs one of the paired images. The pair of images are first encoded in a Siamese manner by the same weight-sharing ViT encoder~\cite{dosovitskiy2020image}, resulting in two token representations, $F_{1}$ and $F_{2}$:
% \noindent \textbf{Robust depth features extraction.}  As shown in Fig.~\ref{fig:architecture}, the DUSt3R component consists of two identical branches, each processing one image and comprising an image encoder, a decoder. Since subsequent operations rely only on the features, components such as the regression head are not required in this process. The two input images are first encoded in a Siamese manner by the same weight-sharing ViT encoder, resulting in two token representations, $F_{1}$ and $F_{2}$ :
\begin{equation}
F_{1} = Encoder(I_{1}), F_2 = Encoder(I_2). \label{dust3r1}
\end{equation}
The decoder then performs joint reasoning on these intermediate features. Through self-attention and cross-attention mechanisms, the decoder effectively processes the view-related information embedded in these features. Specifically, the self-attention module enables the features to focus on other important regions within the current view, while the cross-attention module allows the features to attend to relevant regions across different views. This stacked structure, in which information is continuously exchanged, ensures that both branches acquire cross-view features after decoding. This process is a common technique in many advanced architectures and is crucial for reconstructing an entire scene using image pairs. After decoding, we obtain robust depth features provided by DUSt3R:
\begin{equation}
\begin{aligned}
G_i^1 &= DecoderBlock_i^1(G_{i-1}^1, G_{i-1}^2), \\
G_i^2 &= DecoderBlock_i^2(G_{i-1}^2, G_{i-1}^1),
\end{aligned}
\end{equation}
for \( i = 1, \ldots, B \) for a decoder with \( B \) blocks and initialized with encoder tokens \( G_0^1 := F^1 \) and \( G_0^2 := F^2 \). Here, \( \textit{DecoderBlock}_i^v(G^1_{i-1}, G^2_{i-1}) \) denotes the \( i \)-th block in branch \( v \in \{1, 2\} \), \( G^1_{i-1} \) and \( G^2_{i-1} \) are the input tokens, with \( G^2_{i-1} \) the tokens from the other branch. It should be noted that the parameters of DUSt3R are frozen.

\noindent \textbf{Multi-level features fusion and refinement.}
By processing the features outputted by the multiple decoder blocks, multi-level features are obtained, which are subsequently fused to serve as inputs for downstream tasks. To ensure dimensional consistency, we select the features from the last two scales for fusion within our model. Additionally, we find that utilizing refined features enhances performance. Therefore, we employ a lightweight 2D U-Net~\cite{ronneberger2015u, rombach2022high} to refine the obtained features and incorporate them as residuals into the original feature set. We obtain the refined feature as:
\begin{equation}
\tilde{G} = G + \Delta{G},
\end{equation}
where \( G \) is the original feature and \( \Delta{G} \) is the residual output of the network. At this point, the obtained features \( \tilde{G} \) can be utilized for the subsequent prediction phase. For simplicity, the feature \( \tilde{G} \) is replaced by \( G \) in the following.

\noindent \textbf{Cost volume construction and Gaussian parameters prediction.} It is an intuitive approach to directly use the decoded features to predict the parameters of Gaussian primitives aligned with the pixels. However, the implicit camera representation adopted by DUSt3R limits the implementation of this method, which cannot be directly integrated into the existing training pipeline. Additionally, the depth information provided by DUSt3R lacks scale information and is not sufficiently accurate.  To address these limitations, we employ the same cost volume construction method as MVSplat~\cite{chen2024mvsplat}. The cost volume effectively reflects the degree of matching between features and provides a smoother depth representation~\cite{xu2023unifying, xu2020aanet}. This also enables us to convert the implicit camera model into an explicit model and learn depth in a unified scale space. Specifically, given the feature \(\displaystyle G \in \mathbb{R}^{\frac{H}{4} \times \frac{W}{4} \times C}\) and the corresponding camera pose \(\displaystyle P \in \mathbb{R}^{4 \times 4}\), MVS approach is used to sample \(R\) different depth candidates \(\displaystyle V = \{d_1, d_2, \ldots, d_R\}\) from \(\displaystyle [d_{near}, d_{far}]\). Then, the matching feature \(\displaystyle G^{ij}_{d_k} \in \mathbb{R}^{\frac{H}{4} \times \frac{W}{4} \times C}\) is sampled by warping the \(j\)-th view feature \(G^j\) to the \(i\)-th view according to depth \(d_k\), which is denoted as:
\begin{equation}
G^{ij}_{d_k} = Warp(G^j,\, P^i,\, P^j,\, d_k),
\label{eq:warp}
\end{equation}
where \(Warp\) is the warping and sampling operation.

The matching features from other views are utilized to generate the cost volume feature \(\displaystyle G^i_c = \{G^i_{d_1},\, G^i_{d_2},\, \ldots,\, G^i_{d_R}\}\). It is computed as
\begin{equation}
G^i_{d_k} 
= \frac{1}{N-1} \sum_{j=1,\,j\neq i}^{N} 
  \frac{G^{ij}_{d_k} \cdot G^i}{\sqrt{C}}.
\label{eq:costvolume}
\end{equation}
The depth of the \(i\)-th view \(\displaystyle D^i \in \mathbb{R}^{\frac{H}{4} \times \frac{W}{4}}\) is computed as
\begin{equation}
D^i = \mathrm{softmax}\bigl(G^{i}_{c}\bigr) \, V.
\label{eq:depth}
\end{equation}
The depth will be unprojected to determine the position of the Gaussian parameters. Similar to MVSplat, we use the cost volume \(\displaystyle G^i_c \) in conjunction with the original features \(\displaystyle G \) when predicting Gaussian parameters, except for their position. The key difference lies in how our features are obtained, as they are derived from the vision foundation model. This incorporation significantly enhances the performance of our model compared to current methods.
% Similar to MVSplat, we use the cost volume \(\displaystyle G^i_c \) in conjunction with the original features \(\displaystyle G \) as inputs when predicting Gaussian parameters. The key difference lies in the origin of our features, which are derived from a visual foundation model. This integration significantly enhances the performance of our model compared to previous methods.

% The cost volume features are fed into a lightweight CNN to obtain 
% refined \(\displaystyle \hat{G}^i_{c}\) and the Gaussian feature \(\displaystyle G^{GS}\). 

\subsection{Integrating Semantics for Enhanced Performance}\label{semantic_methods}

The lunar scenes encompass rich semantic information, which can originate from expert annotations or be obtained through segmentation models such as SAM~\cite{kirillov2023segment}. Semantic information serves as a highly effective prior in many tasks \cite{wu2019semantic, ding2023exploring}, inspiring us to design methods that incorporate semantics to improve model performance in lunar reconstruction tasks.

\begin{figure}[htb]
    \vspace{-1em}
    \centering
    \includegraphics[width=\linewidth]{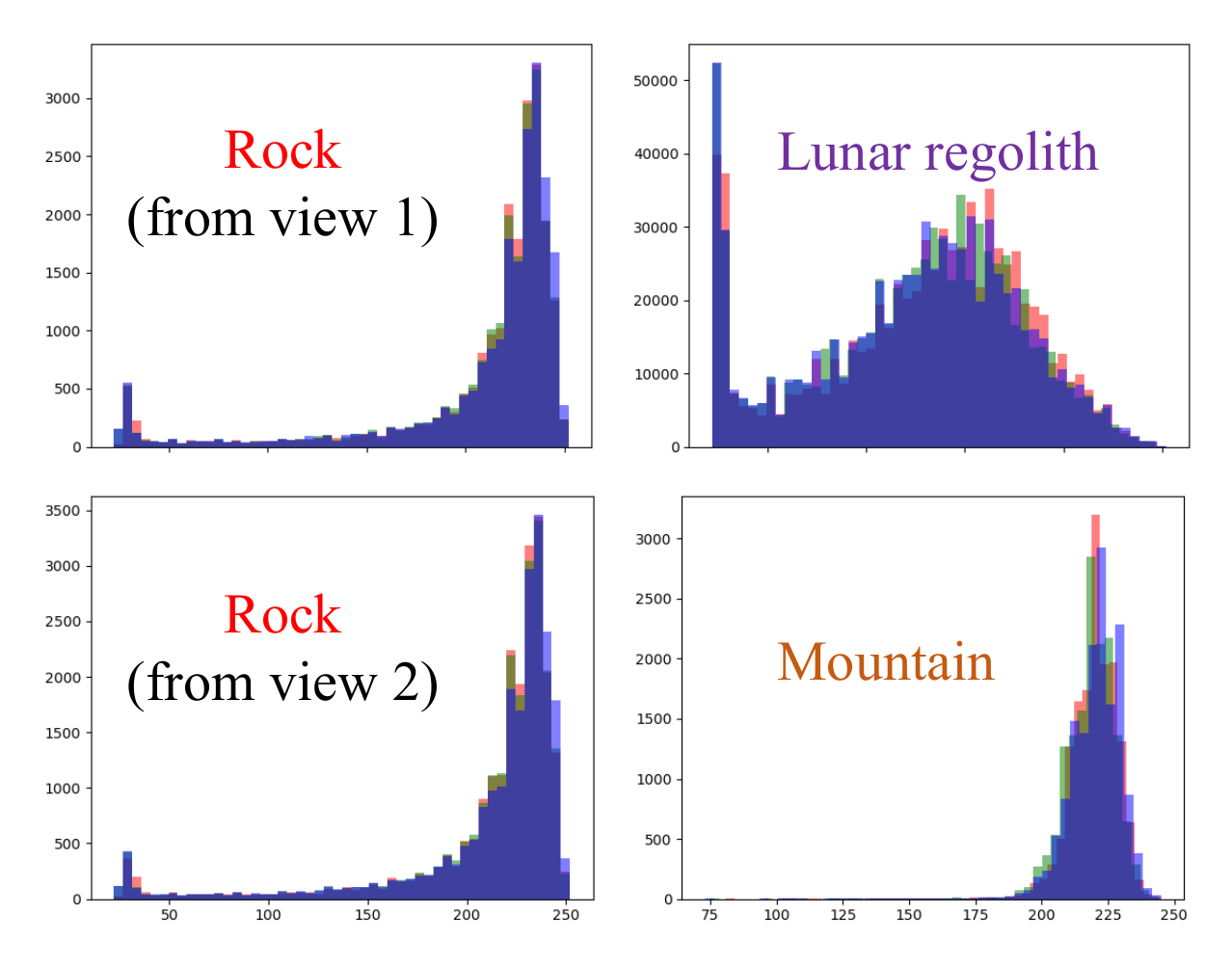}
    \vspace{-13pt}
     \caption{\textbf{Color distribution histograms for semantic categories.} The color histograms demonstrate significant differences in color distribution between specific semantic categories, while the color distributions for the same semantic category show minimal variation. These patterns demonstrate that regions belonging to different semantic categories possess distinct color characteristics, whereas regions within the same semantic category exhibit high similarity. This property allows semantic information to serve as an effective feature for input.}
    \vspace{-12pt}
    \label{fig:semantic_diff}
\end{figure}

\noindent \textbf{Semantic fusion.} Compared to indoor scenes, the semantics of objects in lunar scenes better reflect their characteristics. For instance, different categories of cups typically exhibit markedly distinct texture characteristics. Conversely, rocks of various shapes often possess similar texture features, and rock surfaces differ significantly from those of lunar regolith, which is shown in Fig.\ref{fig:semantic_diff}. Semantic information must also account for certain special cases, such as scenarios where the texture differences between lunar regolith and impact craters are minimal. This inspires us to incorporate semantic information as a prior into the model to improve the accuracy of predicting Gaussian parameters. Specifically, we concatenate the one-hot encoded semantic information with the original features to predict the depth map and Gaussian parameters. When semantics are unreliable or absent, we gate out the fusion branch and use only appearance- and geometry-based features.

\noindent \textbf{Semantic ranking loss.}
In extremely distant regions such as sky and lunar mountains, standard depth estimators often produce imprecise predictions and rarely refine them~\cite{yang2024depth,yang2024depthv2}. Rather than imposing hard background priors like setting background splats to zero opacity or clamping depth to a fixed far plane, which can erase genuine far-range structure and introduce scale bias, we employ a semantic ranking loss inspired by DIW-style supervision~\cite{chen2016single,xian2020structure} to regularize low-texture regions. Specifically, we sample from two categories \(c_1\) and \(c_2\) whose depth relationship is known. Assuming \(c_1\) is nearer than \(c_2\), two pairs of points \([p_{1,c_1}, p_{2,c_2}]\) and \([p_{1,c_2}, p_{2,c_2}]\) can be obtained. We treat the point closer to the camera as a reference depth point, while the more distant point is used for depth optimization.
\begin{equation}
\small
\begin{aligned} \label{eqn:photo_loss}
% \footnotesize
\mathcal{L}_{\text{Ranking}}(p_1 - p_2) =
\begin{cases} 
\log\left(1 + \exp\left(p_1 - p_2\right)\right), & p_1 \in c_1, \\
(p_1 - p_2)^2, & p_1 \in c_2.
\end{cases}
\end{aligned}
\end{equation}

The pairwise ranking loss, where \( p \) is defined in the logarithmic depth space, ensures that background regions remain farther away than the foreground and that depth transitions among background regions become smoother.\
The semantic ranking loss enforces ordinal relations across regions rather than metric depth, so it does not require specifying an absolute far distance. It applies when coarse semantics are available and can be instantiated on any pair of regions with a known farther–nearer relation, not only sky and terrain. The term imposes negligible computational overhead and includes a fallback mechanism that disables the loss when semantics are unavailable or of low quality.

\subsection{Heuristic Resampling Strategy}\label{resample_strategy}

Lunar scenes exhibit a wide depth range and rover motion is predominantly forward. Mission-imposed resource constraints severely limit both the number and diversity of viewpoints, which leaves many surfaces sparsely observed. As a result, distant regions or areas visible from only one viewpoint are hard to reconstruct in detail under a two-view setting. Motivated by these constraints, we introduce a self-supervised heuristic resampling strategy to improve novel-view quality.

\begin{figure}[t]
    % \vspace{-1em}
    \centering
    \includegraphics[width=0.7\linewidth]{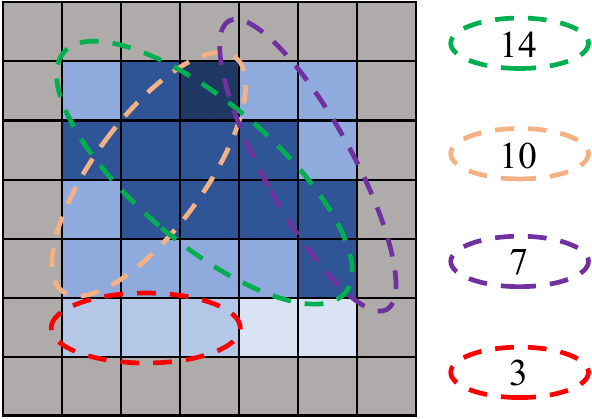}
    % \vspace{-5pt}
    \caption{\textbf{Gaussian coverage statistics.} First, we count the number of valid pixels occupied by each Gaussian. Then, for each pixel, we sum the occupied pixel counts from all Gaussians that cover it. The entire process runs in parallel on the GPU.}
    % \vspace{-15pt}
    \label{fig:resample_strategy_gaussian}
    % \vspace{-15pt}
\end{figure}

After splatting, we compute two view-level statistics for a candidate view as illustrated in Fig.\ref{fig:resample_strategy_gaussian}. Let the image lattice be \(\Omega=\{1,\dots,W\}\times\{1,\dots,H\}\) and \(\mathcal{S}=\{1,\dots,|\mathcal{S}|\}\) index splats.
Rasterization yields each footprint \(M_i^{(v)}\subseteq\Omega\) and its covered-pixel count
\[
A_i^{(v)} \;=\; \bigl|M_i^{(v)}\bigr| ,\qquad i\in\mathcal{S}.
\]
We then accumulate, for every pixel \(x\in\Omega\), the total footprint size of splats that cover it:
\[
T^{(v)}(x)\;=\;\sum_{i\in\mathcal{S}} A_i^{(v)}\,\mathbf{1}\!\left[x\in M_i^{(v)}\right].
\]
Normalizing these totals over the image gives a probability function
\[
p^{(v)}(x)\;=\;\frac{T^{(v)}(x)}{\sum_{y\in\Omega} T^{(v)}(y)}\,,
\]
and the view score is the entropy
\[
H^{(v)}\;=\;-\sum_{x\in\Omega} p^{(v)}(x)\,\log p^{(v)}(x).
\]
Intuitively, \(H^{(v)}\) is large when coverage is spatially dispersed and many footprints overlap, which signals under-coverage or ambiguity in a sparse two-view setting. We therefore use \(H^{(v)}\) as a prioritization score and select candidate locations with the highest values during resampling. This procedure is self-supervised and does not require ground-truth images. Owing to its parallel implementation, the additional computational overhead is negligible. 

\section{Experiments}
In this section, we provide a comprehensive description of the data format and fundamental experimental configurations used in our study, including datasets, evaluation metrics, and implementation details, which are discussed in subsections~\ref{dataset} and~\ref{setting}. These specific details help us to better understand the working parts of the method and allow reproducing the entire experiment. We compared and analyzed the performance of our proposed method against several baseline approaches on lunar datasets, as detailed in subsection~\ref{cp_result}, to illustrate the improvement of our method over state-of-the-art techniques. 
Additionally, we conducted ablation experiments to validate the effectiveness of each component within the overall framework, as described in subsection~\ref{ar}, and further demonstrated the potential to extend our framework to advanced architectures like VGGT~\cite{wang2025vggtvisualgeometrygrounded}.

\subsection{Datasets}\label{dataset}
We used the LuSNAR~\cite{liu2024lusnar} lunar sequence dataset that is compatible with our training pipeline. We also created a MoonBlender dataset with weak textures and more complex rock distributions, and we added qualitative evaluations on Chang’e mission imagery to verify effectiveness on real-world data.

\begin{figure}[t]
    \centering
    \hspace*{0.04\linewidth}\includegraphics[width=\linewidth]{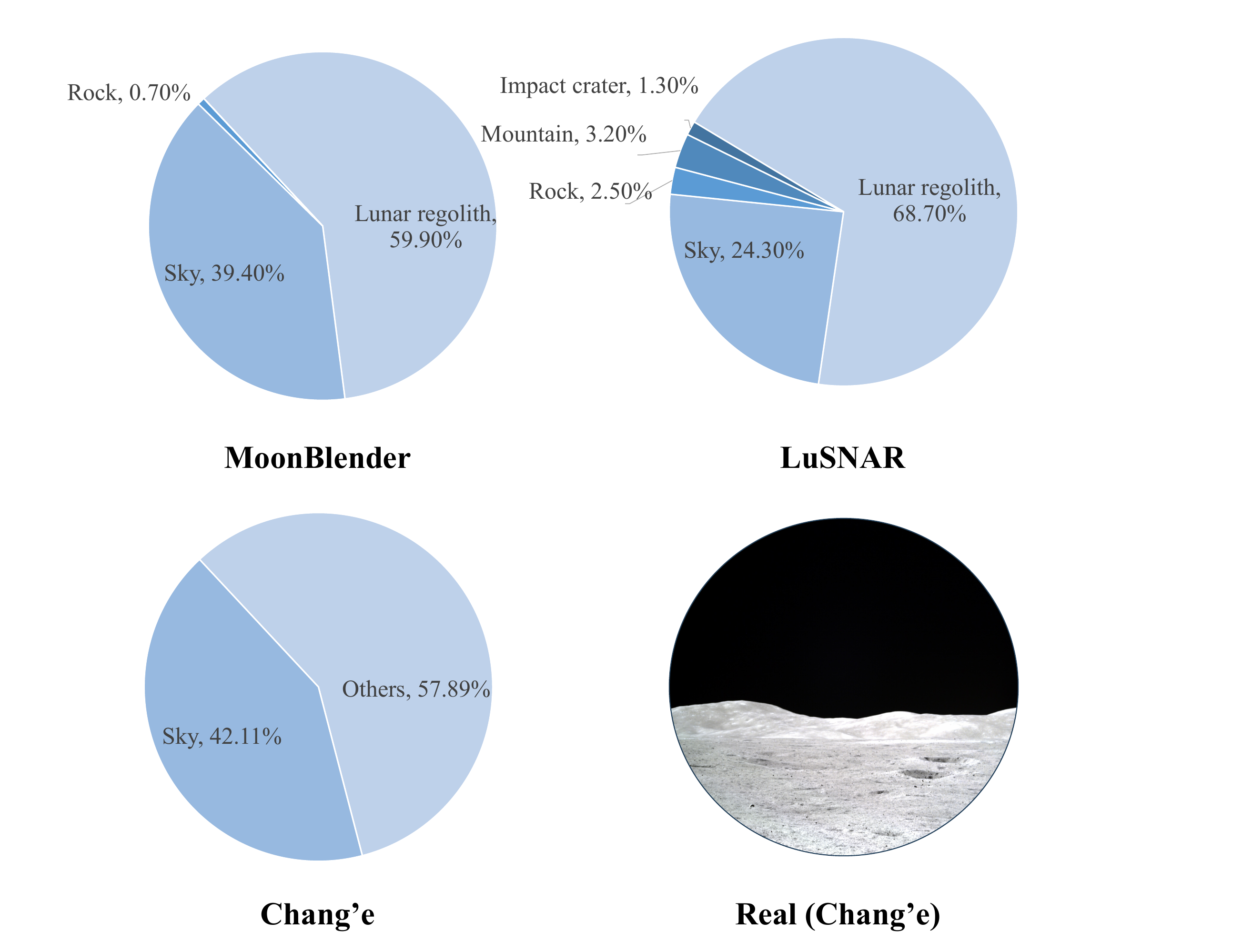}
    \caption{\textbf{Semantic pixel distributions and real-data evidence.} We show LuSNAR and MoonBlender distributions, the sky versus other pixel ratio for Chang’e imagery, and a real Chang’e example. We randomly sampled 96 CE4 TCAM images. 53 images had sky background, and the reported ratios are computed over this subset.}
    \label{fig:percentage}
\end{figure}

\textbf{LuSNAR} is a multi-scene multi-label lunar benchmark dataset built for lunar exploration missions, using Unreal Engine 4~\cite{unrealengine} as a simulator for its project. 
% It has reliable data support for developing and verifying other algorithms. It is a good choice for deep learning models that require a lot of training. 
And for our architecture, it has serialized data, and each image contains real semantic labels and camera poses, which can provide the most basic guarantee for our task. In addition, LuSNAR has multiple scenes, allowing us to verify our algorithm in different scenes. It includes five semantic categories, namely sky, lunar regolith, rocks, impact craters, and mountains. The semantic pixel ratio is shown in Fig.\ref{fig:percentage}. The most important visual sense is that the rocks contained are relatively large irregular rocks that are relatively few. The sky belongs to the weak texture area and has a large pixel ratio.

\textbf{MoonBlender} is a lunar dataset we synthesized using Blender~\cite{blender}, which not only extends the LuSNAR dataset but also serves to validate the problem of weak texture, sparse view reconstruction. 
% \revadd{Such high-proportion weak-texture backgrounds are common in real captures (e.g., in publicly available Chang’e imagery). Given fixed panorama sweep patterns and narrow FOV optics, and operations constrained by illumination/thermal windows, high sky-background frames are objectively hard to avoid~\cite{Yang_2015}. As shown in Fig.\ref{fig:percentage}, we selected a set of real images with prominent sky backgrounds and reported their semantic pixel proportions; these measurements confirm that weak-texture backgrounds are indeed prevalent in real data.} 
Real captures can produce frames with a large sky background. Panorama sweep patterns, narrow field of view optics, and operations constrained by illumination and thermal windows limit pointing range and increase the incidence of large sky background frames~\cite{Yang_2015}. To quantify how often this occurs, we randomly sampled 96 images from the CE4 TCAM dataset, which contains Terrain Camera images captured by the Chang’e 4 lander during surface operations. 43 images had no sky background, and 53 images contained sky background. Among the images with sky background, the mean background pixel ratio was 42.11\% as shown in Fig.\ref{fig:percentage}. These measurements show that frames dominated by sky are common in practice and that the background can occupy a large portion of the image when it occurs.
% Because Gaussian has excellent scene fitting ability, weakly textured data can further evaluate the robustness of existing methods. 
As mentioned above when analyzing the characteristics of LuSNAR data, MoonBlender uses more rocks to simulate a more complex ground environment. At the same time, the sky pixels have a larger proportion and a weaker texture (the value of the pixels is close to 0). This setting helps to evaluate the robustness of different sparse view reconstruction methods for the lunar environment. This is reasonable because the background of the moon has also weak texture when the forward-moving spacecraft collects data in the real environment. Slightly different from LuSNAR, we only have three semantic labels in this dataset: sky, lunar regolith, and rock. Our focus is on using this dataset to validate the effect of weak texture on the model. We referred to a known method for building planetary datasets \cite{liu2023rockformer} and selected five traversable routes. We rendered 1500 images with a resolution of 1200\(\times \)1200 along the virtual cameras set up on these routes. To simulate illumination changes that affect reconstruction, we vary the Sun azimuth and elevation across the dataset. Similarly, this ensures that our data sequence conforms to the training pipeline. Due to Blender's rich interface and excellent graphics processing capability, we can directly obtain metadata such as camera internal and external parameters and semantic segmentation labels of all objects without manual annotation~\cite{cartucho2020visionblender}.

Overall, we used the last 4 of the 9 scenes in LuSNAR as training and validation data. At the same time, all datasets of MoonBlender were used as auxiliary validation. Following the same data settings as previous studies, we trained and evaluated all models on two datasets on images with a resolution of 256\(\times \)256. In addition, we conducted qualitative validation on Chang’e mission imagery to assess performance on real data. Images were grouped by acquisition time, camera poses were estimated with COLMAP for each group, and final view pairs were selected using the same baseline-filtering strategy as pixelSplat.

\begin{table*}[t]
% \centering

% \vspace{5pt}

\caption{
\textbf{Quantitative comparisons.} We evaluated our method by rendering three novel view images from four reference viewpoints for each scene. The performance was determined by averaging across all scenes. The inference time includes both scene encoding and rendering time, tested on a single Q6000 GPU. The number of trainable parameters was computed with pytorch-lightning\cite{Falcon_PyTorch_Lightning_2019}.
}
\resizebox{\textwidth}{!}{

\begin{tabular}{@{}l|ccc|ccc|c|c@{}}
\toprule
\multirow{2}{*}{Methods}                        & \multicolumn{3}{c|}{LuSNAR~\cite{liu2024lusnar}}             & \multicolumn{3}{c|}{MoonBlender}                              & Inference Time        & Trainable Params  \\
                                                & PSNR$\uparrow$    & SSIM$\uparrow$    & LPIPS$\downarrow$     & PSNR$\uparrow$    & SSIM$\uparrow$    & LPIPS$\downarrow$     & (s)                   & (M)               \\ 
\midrule
Du et al.~\cite{du2023learning}                 & 13.26             & 0.347             & 0.727                 & 14.01             & 0.374             & 0.722                 & 2.870                 & -                 \\
MuRF~\cite{xu2024murf}                          & 14.32             & 0.283             & 0.443                 & 14.88             & 0.287             & 0.441                 & -                     & -                 \\
pixelSplat~\cite{charatan2024pixelsplat}        & \underline{18.89} & \underline{0.525} & \underline{0.360}     & \underline{25.26} & \underline{0.689} & \underline{0.166}     & 0.483                 & 118               \\
MVSplat~\cite{chen2024mvsplat}                  & 16.99             & 0.224             & 0.385                 & 25.11             & 0.676             & 0.192                 & \underline{0.323}     & \textbf{12.0}     \\
MoonGS(Ours)                                    & \textbf{23.80}    & \textbf{0.816}    & \textbf{0.134}        & \textbf{26.08}    & \textbf{0.721}    & \textbf{0.139}        & \textbf{0.200}        & \underline{18.1}  \\ 
\bottomrule
\end{tabular}

}

% \vspace{-10pt}
\label{tab:main_result}
\end{table*}

\begin{figure}[h]
    \centering
    \def\svgwidth{0.98\columnwidth}
    \begingroup%
  \makeatletter%
  \providecommand\color[2][]{%
    \errmessage{(Inkscape) Color is used for the text in Inkscape, but the package 'color.sty' is not loaded}%
    \renewcommand\color[2][]{}%
  }%
  \providecommand\transparent[1]{%
    \errmessage{(Inkscape) Transparency is used (non-zero) for the text in Inkscape, but the package 'transparent.sty' is not loaded}%
    \renewcommand\transparent[1]{}%
  }%
  \providecommand\rotatebox[2]{#2}%
  \newcommand*\fsize{\dimexpr\f@size pt\relax}%
  \newcommand*\lineheight[1]{\fontsize{\fsize}{#1\fsize}\selectfont}%
  \ifx\svgwidth\undefined%
    \setlength{\unitlength}{375bp}%
    \ifx\svgscale\undefined%
      \relax%
    \else%
      \setlength{\unitlength}{\unitlength * \real{\svgscale}}%
    \fi%
  \else%
    \setlength{\unitlength}{\svgwidth}%
  \fi%
  \global\let\svgwidth\undefined%
  \global\let\svgscale\undefined%
  \makeatother%
  \begin{picture}(1,0.76490912)%
    \lineheight{0.1}%
    \setlength\tabcolsep{0pt}%
    \centering
    \put(0, -0.113){\includegraphics[width=\unitlength,page=1]{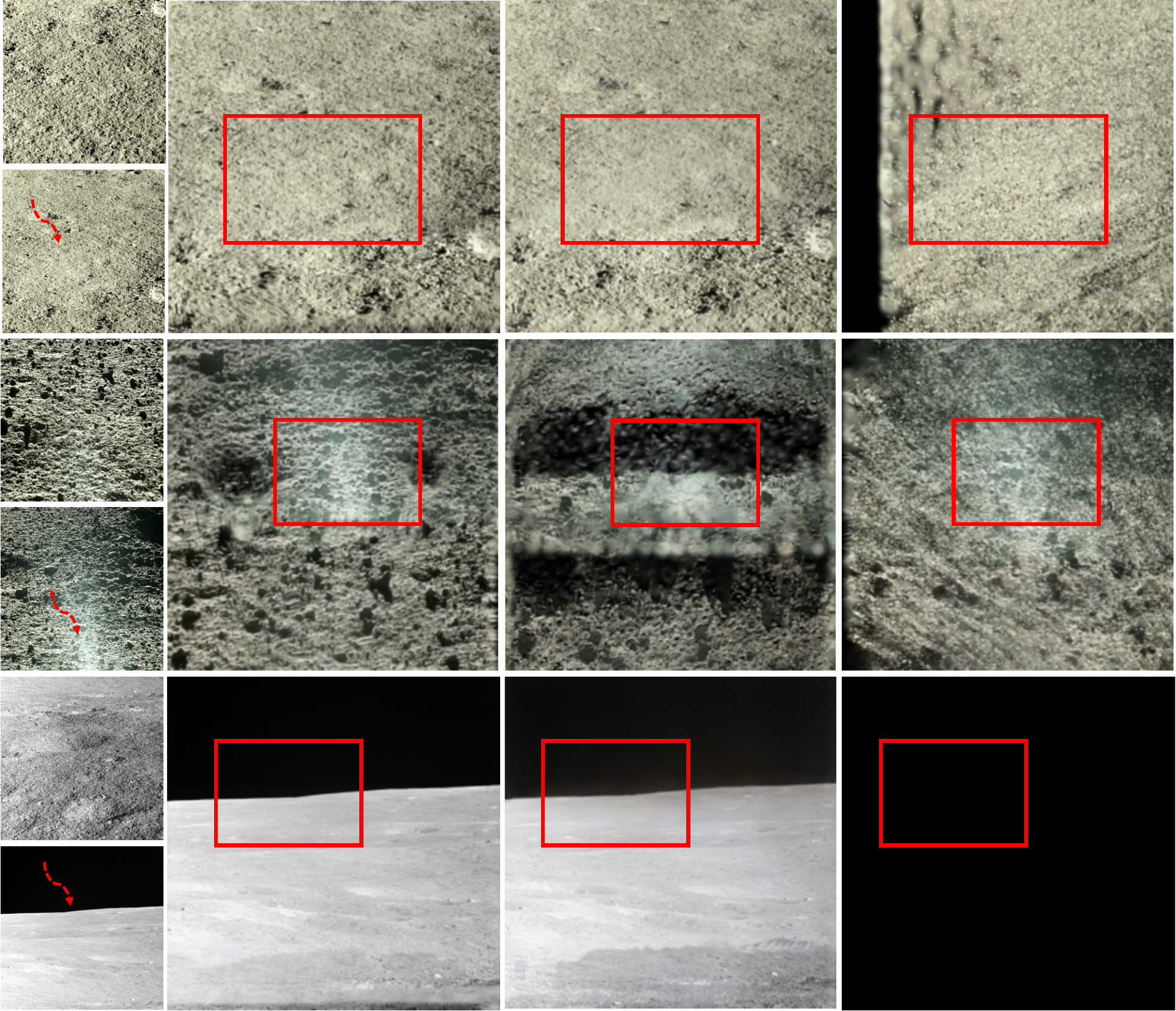}}%
    \put(0.075,0.76490912){\makebox(0,0)[t]{\lineheight{0.1}\smash{\begin{tabular}[t]{c}Ref.\end{tabular}}}}%
    % \put(0.22,0.90320912){\makebox(0,0)[t]{\lineheight{0.1}\smash{\begin{tabular}[t]{c}Target View\end{tabular}}}}%
    \put(0.28,0.76490912){\makebox(0,0)[t]{\lineheight{0.1}\smash{\begin{tabular}[t]{c}Ours\end{tabular}}}}%
    \put(0.575,0.76490912){\makebox(0,0)[t]{\lineheight{0.1}\smash{\begin{tabular}[t]{c}MVSplat~\cite{chen2024mvsplat}\end{tabular}}}}%\textbf{}
    \put(0.855,0.76490912){\makebox(0,0)[t]{\lineheight{0.1}\smash{\begin{tabular}[t]{c}pixelSplat~\cite{charatan2024pixelsplat}\end{tabular}}}}%
  \end{picture}%
\endgroup%

    \vspace{2.6em}
    \caption{\textbf{Qualitative results on real Chang’e imagery.} Given reference image pairs (left column) to render an intermediate viewpoint at the interpolated camera pose. We show three groups (top to bottom) under distinct illumination conditions. MoonGS remains consistent with the input pairs and is more robust to lighting variation, producing fewer artifacts than pixelSplat and MVSplat. PixelSplat exhibits reconstruction failures and noticeable artifacts in challenging lighting, and MVSplat is less robust to illumination changes. Red boxes indicate regions of interest for comparison, and arrows in the reference pair mark the locations of these regions.}
    \label{fig:change_gt}
\end{figure}

\begin{figure*}[t]
    \centering
    
    \def\svgwidth{1\textwidth}
    \begingroup%
  \makeatletter%
  \providecommand\color[2][]{%
    \errmessage{(Inkscape) Color is used for the text in Inkscape, but the package 'color.sty' is not loaded}%
    \renewcommand\color[2][]{}%
  }%
  \providecommand\transparent[1]{%
    \errmessage{(Inkscape) Transparency is used (non-zero) for the text in Inkscape, but the package 'transparent.sty' is not loaded}%
    \renewcommand\transparent[1]{}%
  }%
  \providecommand\rotatebox[2]{#2}%
  \newcommand*\fsize{\dimexpr\f@size pt\relax}%
  \newcommand*\lineheight[1]{\fontsize{\fsize}{#1\fsize}\selectfont}%
  \ifx\svgwidth\undefined%
    \setlength{\unitlength}{375bp}%
    \ifx\svgscale\undefined%
      \relax%
    \else%
      \setlength{\unitlength}{\unitlength * \real{\svgscale}}%
    \fi%
  \else%
    \setlength{\unitlength}{\svgwidth}%
  \fi%
  \global\let\svgwidth\undefined%
  \global\let\svgscale\undefined%
  \makeatother%
  \begin{picture}(1,0.92490912)%
    \lineheight{0.1}%
    \setlength\tabcolsep{0pt}%
    \put(0,0){\includegraphics[width=\unitlength,page=1]{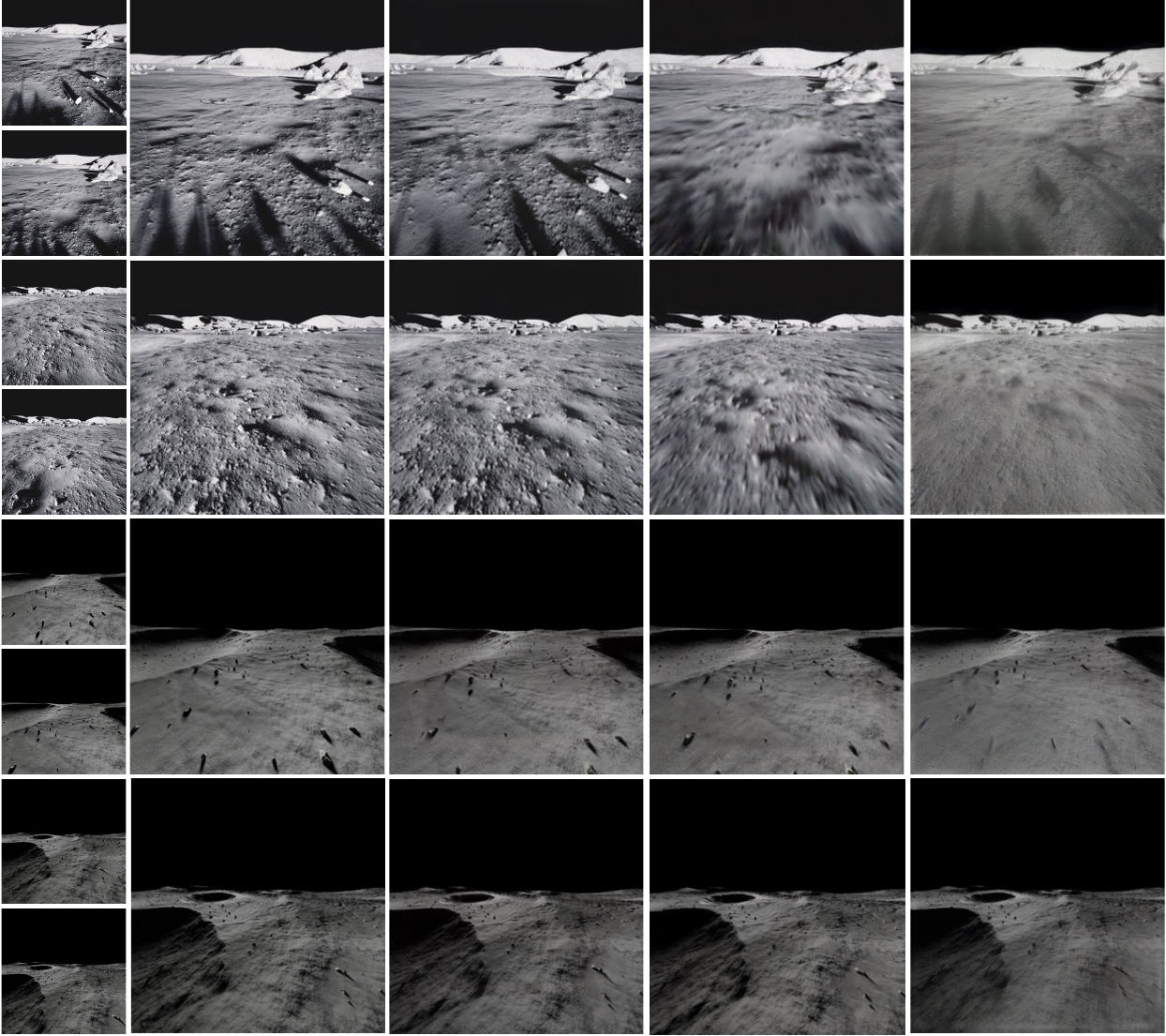}}%
    \put(0.05754546,0.90320912){\makebox(0,0)[t]{\lineheight{0.1}\smash{\begin{tabular}[t]{c}Ref.\end{tabular}}}}%
    \put(0.22,0.90320912){\makebox(0,0)[t]{\lineheight{0.1}\smash{\begin{tabular}[t]{c}Target View\end{tabular}}}}%
    \put(0.44,0.90320912){\makebox(0,0)[t]{\lineheight{0.1}\smash{\begin{tabular}[t]{c}Ours\end{tabular}}}}%
    \put(0.670,0.90320912){\makebox(0,0)[t]{\lineheight{0.1}\smash{\begin{tabular}[t]{c}Pixelsplat~\cite{charatan2024pixelsplat}\end{tabular}}}}%
    \put(0.895,0.90320912){\makebox(0,0)[t]{\lineheight{0.1}\smash{\begin{tabular}[t]{c}MVSplat~\cite{chen2024mvsplat}\end{tabular}}}}%
    % \put(0.90890912,0.91320912){\makebox(0,0)[t]{\lineheight{1.25}\smash{\begin{tabular}[t]{c}pixelNeRF~\cite{pixelnerf}\end{tabular}}}}%
  \end{picture}%
\endgroup%

    % \vspace{-100pt}
    % \includegraphics[width=\linewidth]{figs/main_result/main_result.pdf}
    \caption{\textbf{Qualitative comparison of novel views on the LuSNAR (top) and MoonBlender (bottom) test sets.} Compared to the baseline, our method is able to generate more accurate and perceptually attractive images, especially for lunar regolith close to the camera.}
    % Compared to the baselines, our approach not only produces more accurate and perceptually appealing images, but also generalizes better to out-of-distribution examples like the creek in the bottom row.}
    \label{fig:comparison_joint}
    \vspace{-1em}
    % \vspace{-5pt}
\end{figure*}

\begin{figure*}[t]
    \centering
    
    \def\svgwidth{1\textwidth}
    \begingroup%
  \makeatletter%
  \providecommand\color[2][]{%
    \errmessage{(Inkscape) Color is used for the text in Inkscape, but the package 'color.sty' is not loaded}%
    \renewcommand\color[2][]{}%
  }%
  \providecommand\transparent[1]{%
    \errmessage{(Inkscape) Transparency is used (non-zero) for the text in Inkscape, but the package 'transparent.sty' is not loaded}%
    \renewcommand\transparent[1]{}%
  }%
  \providecommand\rotatebox[2]{#2}%
  \newcommand*\fsize{\dimexpr\f@size pt\relax}%
  \newcommand*\lineheight[1]{\fontsize{\fsize}{#1\fsize}\selectfont}%
  \ifx\svgwidth\undefined%
    \setlength{\unitlength}{375bp}%
    \ifx\svgscale\undefined%
      \relax%
    \else%
      \setlength{\unitlength}{\unitlength * \real{\svgscale}}%
    \fi%
  \else%
    \setlength{\unitlength}{\svgwidth}%
  \fi%
  \global\let\svgwidth\undefined%
  \global\let\svgscale\undefined%
  \makeatother%
  \begin{picture}(1,0.92490912)%
    \lineheight{0.1}%
    \setlength\tabcolsep{0pt}%
    \put(0,0.1435){\includegraphics[width=\unitlength,page=1]{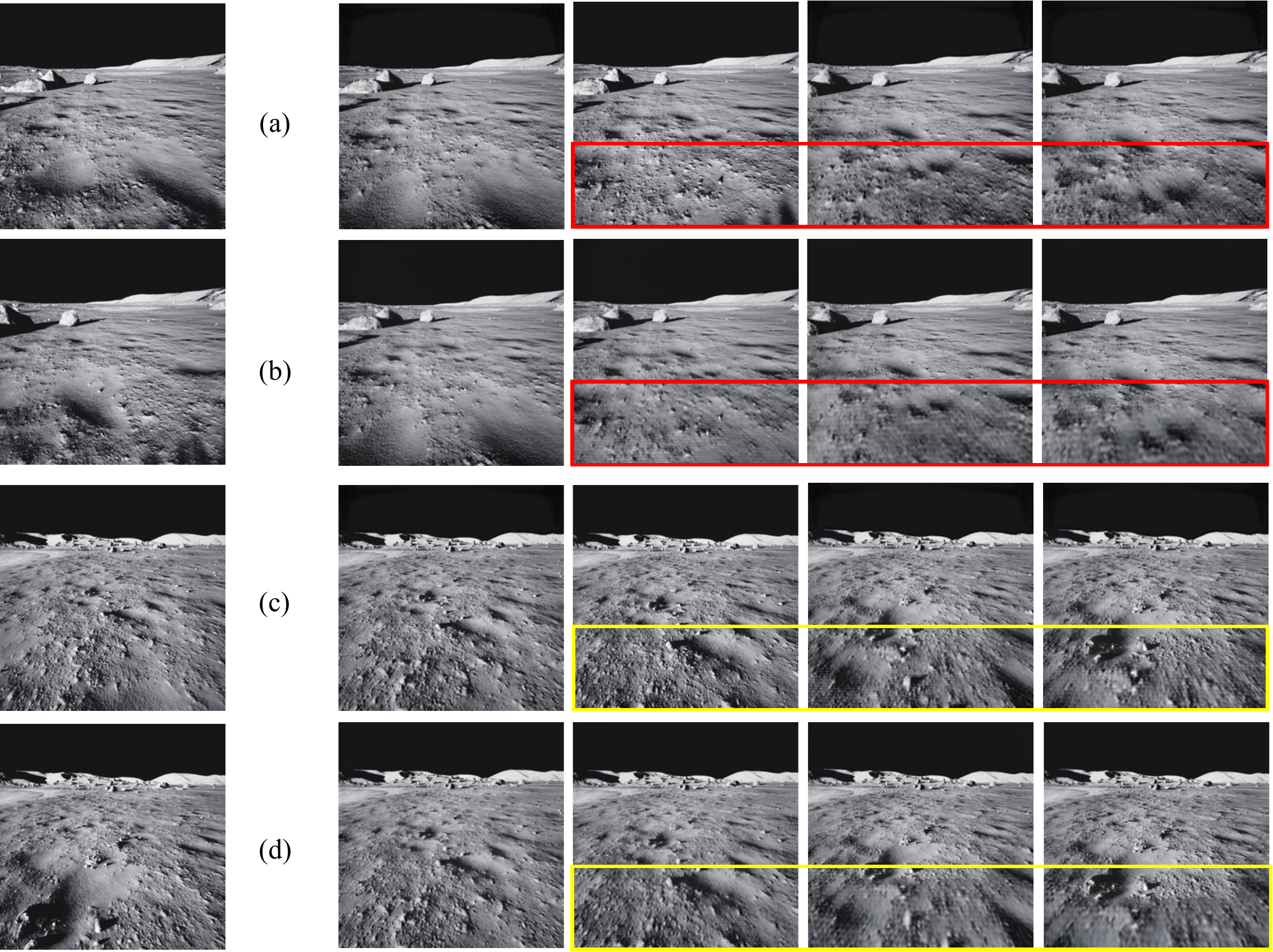}}%
    \put(0.085,0.90320912){\makebox(0,0)[t]{\lineheight{0.1}\smash{\begin{tabular}[t]{c}Ref.\end{tabular}}}}%
    \put(0.36,0.90320912){\makebox(0,0)[t]{\lineheight{0.1}\smash{\begin{tabular}[t]{c}Position\textsuperscript{1}\end{tabular}}}}%
    \put(0.54,0.90320912){\makebox(0,0)[t]{\lineheight{0.1}\smash{\begin{tabular}[t]{c}\textbf{Position}\textbf{\textsuperscript{2}}\end{tabular}}}}%
    \put(0.72,0.90320912){\makebox(0,0)[t]{\lineheight{0.1}\smash{\begin{tabular}[t]{c}Position\textsuperscript{3}\end{tabular}}}}%
    \put(0.905,0.90320912){\makebox(0,0)[t]{\lineheight{0.1}\smash{\begin{tabular}[t]{c}Position\textsuperscript{4}\end{tabular}}}}%
    % \put(0.90890912,0.91320912){\makebox(0,0)[t]{\lineheight{1.25}\smash{\begin{tabular}[t]{c}pixelNeRF~\cite{pixelnerf}\end{tabular}}}}%
  \end{picture}%
\endgroup%

    \vspace{-7em}
    % \includegraphics[width=\linewidth]{figs/main_result/main_result.pdf}
    % \caption{\textbf{Qualitative comparison of novel views on the LuSNAR (top) and MoonBlender (bottom) test sets.} Compared to the baseline, our method is able to generate more accurate and perceptually attractive images, especially for lunar regolith close to the camera.}
    \caption{\textbf{Comparison of rendering results with and without heuristic resampling strategy.} We present two groups of rendering results across four viewpoints: (a, c) with the heuristic resampling strategy, and (b, d) without the heuristic resampling strategy. The resampled region (\textbf{Position}\textsuperscript{\textbf{2}}) is the selected resampling location. The results demonstrate that heuristic resampling effectively reduces artifacts and produces a more realistic textural appearance.}
    % Compared to the baselines, our approach not only produces more accurate and perceptually appealing images, but also generalizes better to out-of-distribution examples like the creek in the bottom row.}
    \label{fig:resample_compare}
    % \vspace{-5pt}
\end{figure*}

\subsection{Experimental Settings} \label{setting}

\noindent \textbf{Comparison methods.} We compared methods for feed-forward synthesis of novel views from sparse views, including NeRF and 3DGS. Du et al.~\cite{du2023learning} is a light field-based rendering method that can aggregate epipolar features from the Transformer structure. MuRF~\cite{xu2024murf} is a sparse view reconstruction method based on NeRF. PixelSplat~\cite{charatan2024pixelsplat} and MVSplat~\cite{chen2024mvsplat} are advanced feed-forward methods based on 3DGS. Our method is mainly improved from MVSplat. These methods were compared qualitatively and quantitatively with ours. 

\noindent \textbf{Evaluation metrics and training losses.} For image rendering quality, we used standard image quality metrics to quantitatively evaluate and compare our method with other methods: pixel-level PSNR, patch-level SSIM~\cite{wang2004image}, and feature-level LPIPS~\cite{zhang2018unreasonable}. Since our training was trained and evaluated through the standard explicit camera model training pipeline, we could directly compare with the real ground truth. In addition, to more clearly illustrate the change in the overall resource consumption of the model after using the vision foundation model, we evaluated our resource consumption such as the number of trainable parameters and inference time to illustrate the increased resource consumption compared with other models. In addition, we also qualitatively evaluated the accuracy of the depth estimation of the model. It is common and effective to use MSE and LPIPS loss functions with weights 1 and 0.05, respectively during the training stage. Ranking loss with weight 0.01 was used in our study to regularize weakly textured regions. In general, the loss can be described by the following formula:
\begin{equation}
\mathcal{L_{\text{total}}} = \mathcal{L}_{\text{MSE}} + 0.05\times\mathcal{L}_{\text{LPIPS}} + 0.01\times\mathcal{L}_{\text{Ranking}}.
\end{equation}

\noindent \textbf{Hardware and implementation details.}
All timings were measured on NVIDIA Quadro RTX 6000 (Q6000) GPUs. Training used 4 GPUs with a global batch size of 16 (4 images per GPU) at an image resolution of \(256\times256\). We trained for 30{,}000 steps. We then fine-tuned for 500 steps with the ranking loss enabled to better regularize weak textured regions. Evaluation and inference were performed on a single Q6000. Unless otherwise stated, inference timings correspond to a batch size of 1. We optimized all models with Adam using a learning rate of \(1\times10^{-4}\).

\subsection{Comparative Results}\label{cp_result}
\noindent \textbf{Image quality comparison.} We report the quantitative results of our method and the baselines on the LuSNAR and MoonBlender datasets in Tab.~\ref{tab:main_result}. The baseline results were obtained by retraining the original implementations on these datasets. On the LuSNAR dataset, our method demonstrated ability to recover more realistic ground textures. Specifically, while the highest PSNR among the other methods was 18.89 dB, our approach achieved 23.80 dB, indicating a substantial improvement. Moreover, our method outperformed the baselines in terms of SSIM and LPIPS. On the MoonBlender dataset, our method improved the best PSNR (25.26 dB) from the pixelSplat baseline to 26.08 dB, while also outperforming existing baselines on other metrics. Although this dataset is not specifically designed to evaluate reconstruction quality due to its relatively simple visual and environmental settings, our results suggest that the underlying DUSt3R~\cite{wang2024dust3r} vision foundation model enhances the robustness of our approach, enabling it to handle scenes with varying complexity. Overall, our approach combines the structure extraction capability of the vision foundation model DUSt3R and the texture prediction capability of MVSplat to obtain realistic Lunar landscapes.

\begin{figure}[h]
    \centering
    \def\svgwidth{0.98\columnwidth} % 或者 \linewidth
    \begingroup%
  \makeatletter%
  \providecommand\color[2][]{%
    \errmessage{(Inkscape) Color is used for the text in Inkscape, but the package 'color.sty' is not loaded}%
    \renewcommand\color[2][]{}%
  }%
  \providecommand\transparent[1]{%
    \errmessage{(Inkscape) Transparency is used (non-zero) for the text in Inkscape, but the package 'transparent.sty' is not loaded}%
    \renewcommand\transparent[1]{}%
  }%
  \providecommand\rotatebox[2]{#2}%
  \newcommand*\fsize{\dimexpr\f@size pt\relax}%
  \newcommand*\lineheight[1]{\fontsize{\fsize}{#1\fsize}\selectfont}%
  \ifx\svgwidth\undefined%
    \setlength{\unitlength}{375bp}%
    \ifx\svgscale\undefined%
      \relax%
    \else%
      \setlength{\unitlength}{\unitlength * \real{\svgscale}}%
    \fi%
  \else%
    \setlength{\unitlength}{\svgwidth}%
  \fi%
  \global\let\svgwidth\undefined%
  \global\let\svgscale\undefined%
  \makeatother%
  \begin{picture}(1,0.4)%
    \lineheight{0.1}%
    \setlength\tabcolsep{0pt}%
    \centering
    \put(0, -0.113){\includegraphics[width=\unitlength,page=1]{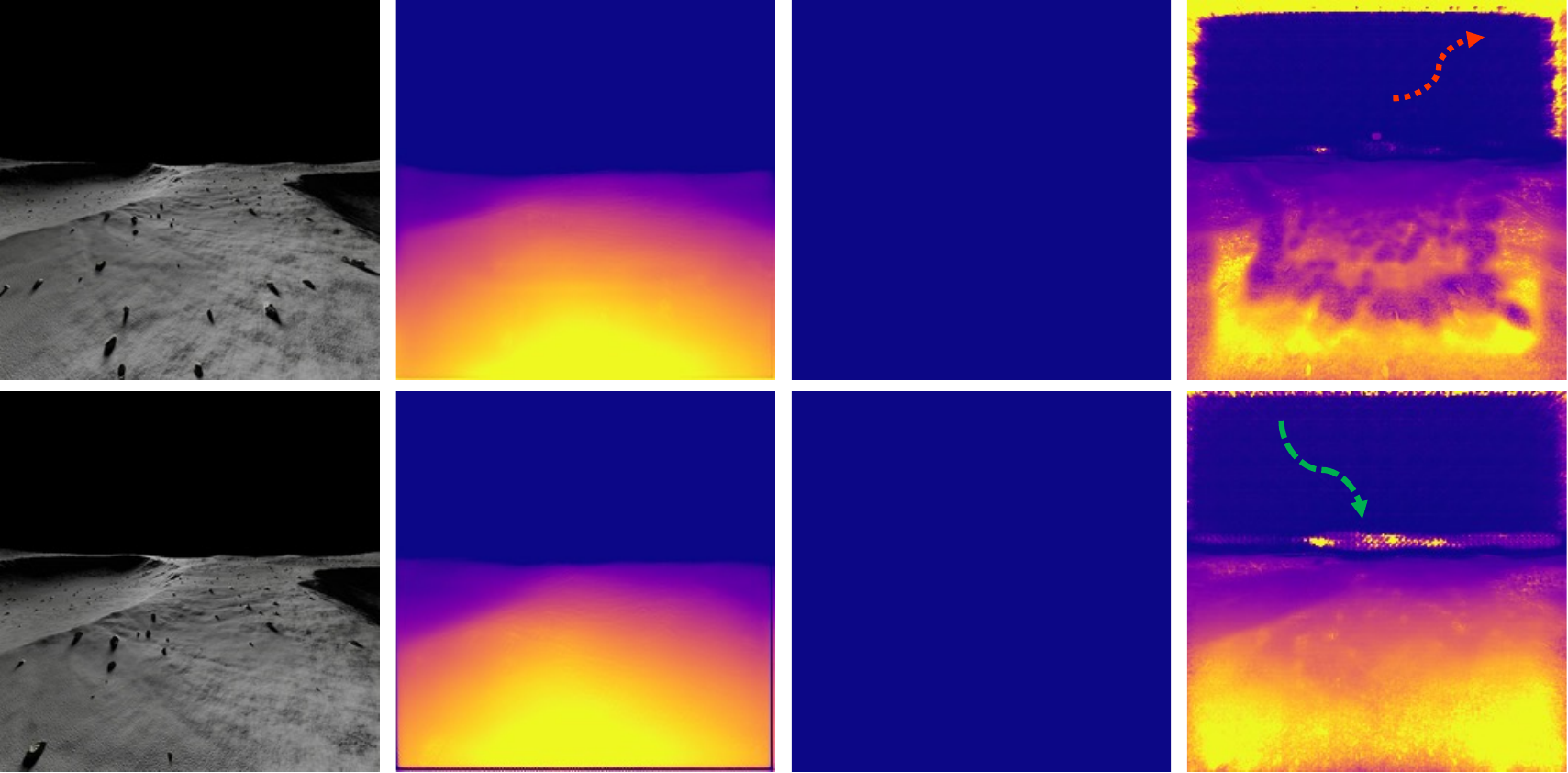}}%
    \put(0.12,0.4){\makebox(0,0)[t]{\lineheight{0.1}\smash{\footnotesize\begin{tabular}[t]{c}Ref.\end{tabular}}}}%
    % \put(0.22,0.90320912){\makebox(0,0)[t]{\lineheight{0.1}\smash{\begin{tabular}[t]{c}Target View\end{tabular}}}}%
    \put(0.37,0.4){\makebox(0,0)[t]{\lineheight{0.1}\smash{\footnotesize\begin{tabular}[t]{c}Ours\end{tabular}}}}%
    \put(0.63,0.4){\makebox(0,0)[t]{\lineheight{0.1}\smash{\footnotesize\begin{tabular}[t]{c}MVSplat~\cite{chen2024mvsplat}\end{tabular}}}}%\textbf{}
    \put(0.88,0.4){\makebox(0,0)[t]{\lineheight{0.1}\smash{\footnotesize\begin{tabular}[t]{c}pixelSplat~\cite{charatan2024pixelsplat}\end{tabular}}}}%

    % \put(0.90890912,0.91320912){\makebox(0,0)[t]{\lineheight{1.25}\smash{\begin{tabular}[t]{c}pixelNeRF~\cite{pixelnerf}\end{tabular}}}}%
  \end{picture}%
\endgroup%

    \textbf{\vspace{2.9em}}
    \caption{\textbf{Qualitative comparison of depth maps on the MoonBlender dataset.} This figure compares depth maps produced by several models on the MoonBlender dataset. Our proposed method (MoonGS) generates both accurate and smoothly varying depth maps. In contrast, MVSplat converges to erroneous depths, and although pixelSplat produces relatively accurate depths, it exhibits noticeable artifacts—such as non-smooth transitions (indicated by the arrows). These findings highlight the importance of robust depth estimation techniques for lunar scene reconstruction.}
    \label{fig:depth_compare}
\end{figure}

Qualitative comparisons in Fig.\ref{fig:comparison_joint} further confirm that our method produced fewer artifacts than the best-performing baseline, pixelSplat, especially under more complex ground conditions. Additionally, when the rover moves forward on the lunar surface, non-overlapping regions near the camera exhibited fewer artifacts in our reconstruction. Notably, while pixelSplat did not perform as well as MVSplat in indoor scenarios, it remained highly competitive in lunar environments. Our method achieved superior reconstruction in challenging areas primarily because the vision foundation model provides more effective depth-related features, laying a stronger groundwork for subsequent depth optimization. Quantitative analyses in Fig.\ref{fig:depth_compare}, reveal that our unprojection-based depth estimations were more reliable compared to other approaches. In contrast, pixelSplat produced highly noisy depth estimates in low-texture regions, and MVSplat often suffered from projection errors when trained on the sparse data of MoonBlender, resulting in Gaussian distributions collapsing to the same depth and leading to incorrect depth estimates. By leveraging robust depth features from the vision foundation model, our method generated more accurate depth estimates during subsequent optimization steps. The depth comparison also demonstrated that our depth maps are not only more reliable but also smoother. Furthermore, incorporating a semantic-based depth ranking loss allowed us to better optimize weakly textured sky regions, enhancing performance in these challenging areas. We see in Fig.\ref{fig:depth_compare} that pixelSplat without regularization produced some incorrect depth estimates in weakly textured sky. This is mainly because the photometric loss is difficult to use for supervising the depth in weakly textured areas. We will continue to analyze this in ablation experiments.

Qualitative results on real Chang’e mission imagery further supported these findings, as shown in Fig.~\ref{fig:change_gt}. We evaluated three distinct illumination conditions and ran all methods on the same image pairs. PixelSplat frequently exhibited reconstruction failures under harsh lighting and produced pronounced artifacts near high-contrast edges. MVSplat was not robust to illumination shifts, and its renderings deviated from the reference views with noticeable texture washout and contrast drift. In contrast, our method adhered closely to the input image pair and showed no reconstruction failures across the tested illumination conditions. This demonstrated applicability to real lunar data and robustness to lighting variation. When illumination differed markedly between the two inputs, a visible boundary could appear. Even so, reconstructions from our method remained closer to the references and preserved fine-grained details.

\noindent \textbf{Resource consumption comparison.} To illustrate the advantages of our approach in terms of resource consumption, we examined several critical aspects. First, regarding inference time, as shown in Tab.~\ref{tab:main_result}, methods based on 3DGS exhibited a significant lead over NeRF-based approaches, and our method outperformed MVSplat—the closest competitor in terms of inference speed—by at least 0.1 seconds. Second, in terms of trainable parameters, pixelSplat contains a substantial number of parameters due to its complex epipolar attention mechanism, while our method only adds a relatively modest number of parameters (around 6M) to MVSplat by incorporating a multi-scale feature fusion U-Net and a semantic fusion module. Nevertheless, we observed a marked performance improvement over MVSplat. Additionally, both modules employed lightweight 2D convolution structures, which enabled superior performance even with fewer overall parameters. This indicates that the performance gains primarily stem from more effective feature representations rather than merely increasing the model size. We kept the DUSt3R backbone frozen and trained only the downstream modules. This reduced the number of trainable parameters and lowered training cost while yielding competitive results in our setting. The choice was consistent with prior work that leveraged frozen visual backbones in low-data regimes~\cite{LUO2025126}. Freezing was also commonly used to mitigate overfitting when data were limited.

Finally, with respect to training steps, our method required only around 30,000 steps yet still achieved better results than the baseline methods, underscoring its lower resource consumption. Overall, considering inference time, the number of trainable parameters, and the required training steps, our method demonstrated clear advantages in resource consumption over the baseline methods.

\noindent \textbf{Image quality comparison with heuristic resampling strategy.} The vision foundation model and semantic ranking loss effectively tackle two key challenges in lunar surface reconstruction: invalid epipolar constraints and unreliable depth estimation in areas with weak texture. The heuristic resampling method aims to address the third problem, the lack of long-distance observations during the forward sampling process of the lunar rover. Fig.\ref{fig:resample_compare} illustrates the rendered images on the LuSNAR dataset both with and without heuristic resampling. We computed statistics in parallel during the rendering process and used a heuristic approach to select the resampled locations. From a visual perspective, incorporating heuristic resampling lead to significantly richer detail and better outcomes. Compared with using image pairs, resampling can alleviate the problem of data sparsity from the data level. Qualitative observations further indicated that these improvements became more pronounced in areas closer to the position of the second reference image. Throughout the process, we did not use the true value, but adopted a self-supervised approach for guidance. This reflected the speed and quality advantages of our method.

\begin{table}[t]
\centering
\small
\vspace{-1em}
% \caption{\textbf{Scalability analysis on the LuSNAR dataset.} The results demonstrate that our framework consistently yields improvements over the baselines, verifying its extensibility and ability to leverage advanced architectures.}
\caption{\textbf{Applicability analysis on the LuSNAR dataset.} Evaluated at a resolution of $224 \times 224$, the results show that our method effectively capitalizes on the VGGT backbone to outperform baselines, verifying its robust applicability and capacity to evolve with advanced foundation models.}
\begin{tabular}{l|ccc}
\toprule
Methods & PSNR $\uparrow$ & SSIM $\uparrow$ & LPIPS $\downarrow$ \\
\midrule
{pixelSplat}~\cite{charatan2024pixelsplat}     & 19.22 & 0.632 & 0.355 \\
{MVSplat}~\cite{chen2024mvsplat}     & 17.80 & 0.542 & 0.377 \\
MoonGS + DUSt3R~\cite{wang2024dust3r} & \underline{23.91} & \underline{0.823} & \underline{0.130} \\
MoonGS + VGGT~\cite{wang2025vggtvisualgeometrygrounded} & \textbf{24.04} & \textbf{0.830} & \textbf{0.128} \\
\bottomrule
\end{tabular}
% \vspace{-5pt}
\label{tab:vggt}
\end{table}

\begin{figure}[!htbp]
    \centering
    \def\svgwidth{\columnwidth} % 或者 \linewidth
    \begingroup%
  \makeatletter%
  \providecommand\color[2][]{%
    \errmessage{(Inkscape) Color is used for the text in Inkscape, but the package 'color.sty' is not loaded}%
    \renewcommand\color[2][]{}%
  }%
  \providecommand\transparent[1]{%
    \errmessage{(Inkscape) Transparency is used (non-zero) for the text in Inkscape, but the package 'transparent.sty' is not loaded}%
    \renewcommand\transparent[1]{}%
  }%
  \providecommand\rotatebox[2]{#2}%
  \newcommand*\fsize{\dimexpr\f@size pt\relax}%
  \newcommand*\lineheight[1]{\fontsize{\fsize}{#1\fsize}\selectfont}%
  \ifx\svgwidth\undefined%
    \setlength{\unitlength}{375bp}%
    \ifx\svgscale\undefined%
      \relax%
    \else%
      \setlength{\unitlength}{\unitlength * \real{\svgscale}}%
    \fi%
  \else%
    \setlength{\unitlength}{\svgwidth}%
  \fi%
  \global\let\svgwidth\undefined%
  \global\let\svgscale\undefined%
  \makeatother%
  \begin{picture}(1,0.4)%
    \lineheight{0.1}%
    \setlength\tabcolsep{0pt}%
    \centering
    \put(0, 0.13){\includegraphics[width=\unitlength,page=1]{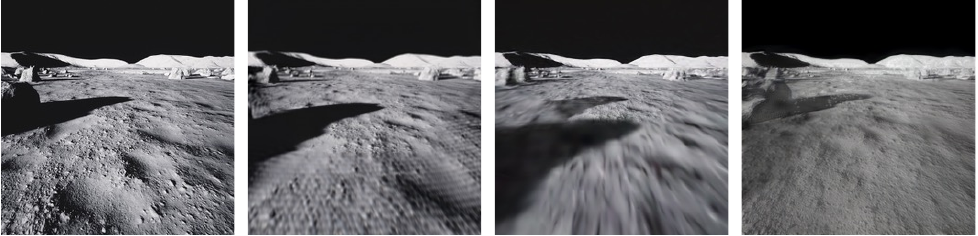}}%
    \put(0.12,0.4){\makebox(0,0)[t]{\lineheight{0.1}\smash{\footnotesize\begin{tabular}[t]{c}Ref.\end{tabular}}}}%
    % \put(0.22,0.90320912){\makebox(0,0)[t]{\lineheight{0.1}\smash{\begin{tabular}[t]{c}Target View\end{tabular}}}}%
    \put(0.37,0.4){\makebox(0,0)[t]{\lineheight{0.1}\smash{\footnotesize\begin{tabular}[t]{c}Ours\end{tabular}}}}%
    \put(0.63,0.4){\makebox(0,0)[t]{\lineheight{0.1}\smash{\footnotesize\begin{tabular}[t]{c}pixelSplat~\cite{charatan2024pixelsplat}\end{tabular}}}}%\textbf{}
    \put(0.88,0.4){\makebox(0,0)[t]{\lineheight{0.1}\smash{\footnotesize\begin{tabular}[t]{c}MVSplat~\cite{chen2024mvsplat}\end{tabular}}}}%

    % \put(0.90890912,0.91320912){\makebox(0,0)[t]{\lineheight{1.25}\smash{\begin{tabular}[t]{c}pixelNeRF~\cite{pixelnerf}\end{tabular}}}}%
  \end{picture}%
\endgroup%

    \textbf{\vspace{-4.6em}}
    \caption{\textbf{Qualitative comparison on the LUSNAR dataset.} We evaluate the effectiveness of integrating the VGGT foundation model within our proposed framework. In terms of visual quality, our method outperforms the baseline model.}
    \label{fig:vggt}
\end{figure}

\noindent \textbf{Applicability to advanced backbones.} 
To demonstrate that our framework is backbone-agnostic and can directly benefit from the evolution of vision foundation models, we replaced the default feature extractor with the state-of-the-art VGGT~\cite{wang2025vggtvisualgeometrygrounded}. 
Rather than a mere compatibility check, this experiment validates the broad applicability of our method. 
By integrating the stronger priors from VGGT while keeping the foundation model frozen (maintaining only 50M trainable parameters), our framework achieves immediate performance gains (Tab.~\ref{tab:vggt}). 
Fig.~\ref{fig:vggt} further illustrates that the VGGT-equipped model significantly outperforms baselines without requiring auxiliary modules. 
This confirms that MoonGS is not limited to a specific encoder but serves as a versatile platform capable of evolving with future advancements in feature representation.

\begin{figure*}[ht]
\centering
\includegraphics[width=\textwidth]{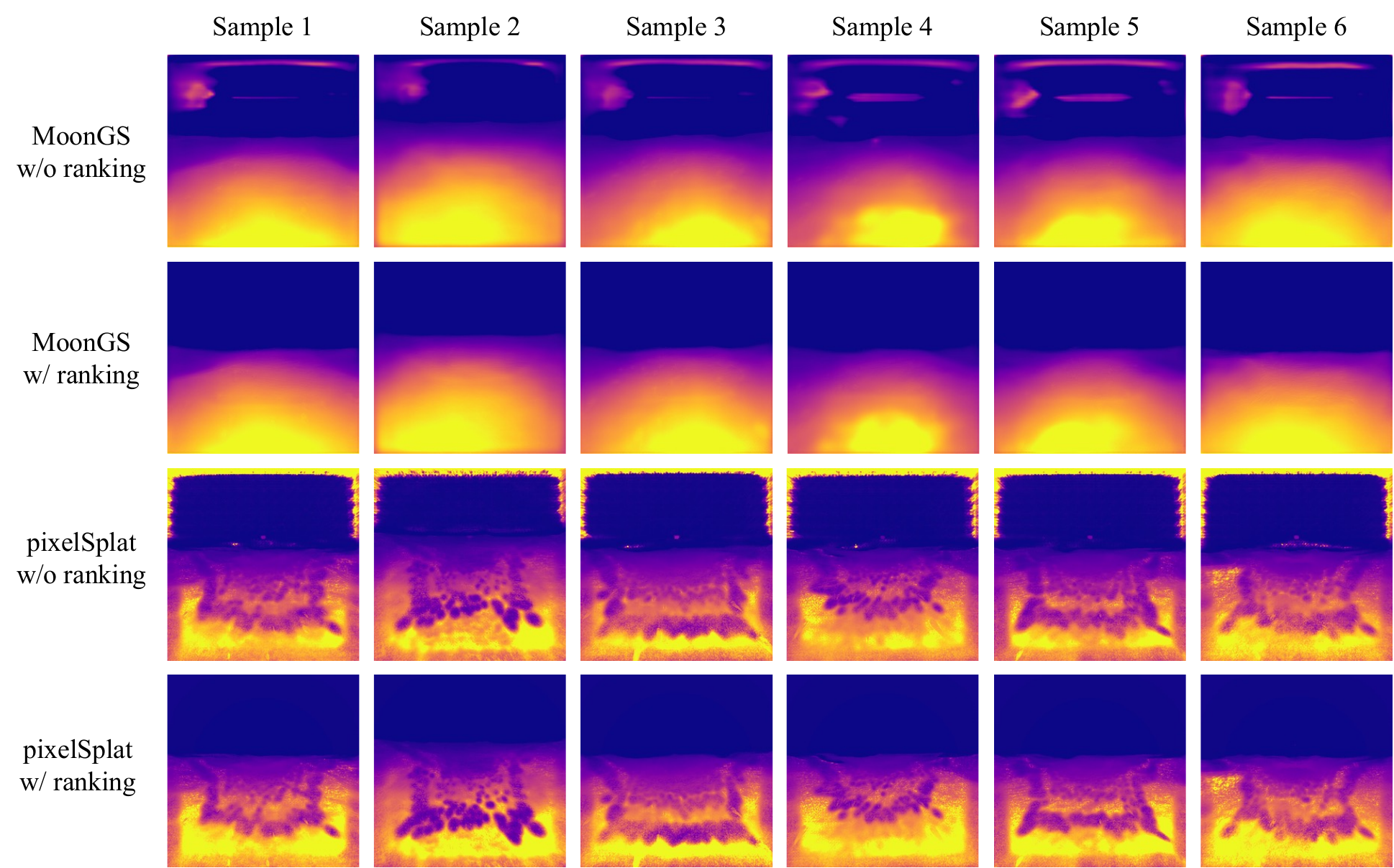}
\caption{\textbf{Impact of ranking loss on depth estimation in weakly textured background areas.} The results show the effect of ranking loss on improving depth estimation in regions with weak texture. In the absence of ranking loss, depth estimates in these areas are often imprecise due to the lack of texture. However, by introducing ranking loss, the model is able to refine the depth predictions in these challenging regions, leading to more reliable results. Moreover, integrating ranking loss with the pixelSplat model yields further improvements in depth accuracy. These results underscore the value of ranking loss in addressing depth estimation issues in weakly textured regions.}
\label{fig:depthloss}
\vspace{-1em}
\end{figure*}

\subsection{Ablation Results}\label{ar}

In this subsection, we systematically examine the effectiveness of our proposed method from several key perspectives. First, we analyzed the benefits of multi-level feature fusion combined with U-Net, which plays a critical role in efficient information extraction and enhancement. Next, we validated the contribution of integrating semantic information. Additionally, we evaluated the semantic ranking loss, demonstrating its effectiveness in further regularizing depth predictions. Our experiments also confirmed that the heuristic resampling strategy significantly enhanced rendering quality. Finally, we integrated our proposed post-processing module into the pixelSplat approach to demonstrate the effectiveness and broader applicability of our method.

\begin{figure}[t]
\centering
\small
    % \vspace{-5pt}
    % \includegraphics[width=\linewidth]{figs/unet/fu.pdf}
    % \vspace{1pt}
    \centering
    \def\svgwidth{0.98\columnwidth} % 或者 \linewidth
    \begingroup%
  \makeatletter%
  \providecommand\color[2][]{%
    \errmessage{(Inkscape) Color is used for the text in Inkscape, but the package 'color.sty' is not loaded}%
    \renewcommand\color[2][]{}%
  }%
  \providecommand\transparent[1]{%
    \errmessage{(Inkscape) Transparency is used (non-zero) for the text in Inkscape, but the package 'transparent.sty' is not loaded}%
    \renewcommand\transparent[1]{}%
  }%
  \providecommand\rotatebox[2]{#2}%
  \newcommand*\fsize{\dimexpr\f@size pt\relax}%
  \newcommand*\lineheight[1]{\fontsize{\fsize}{#1\fsize}\selectfont}%
  \ifx\svgwidth\undefined%
    \setlength{\unitlength}{375bp}%
    \ifx\svgscale\undefined%
      \relax%
    \else%
      \setlength{\unitlength}{\unitlength * \real{\svgscale}}%
    \fi%
  \else%
    \setlength{\unitlength}{\svgwidth}%
  \fi%
  \global\let\svgwidth\undefined%
  \global\let\svgscale\undefined%
  \makeatother%
  \begin{picture}(1,0.4)%
    \lineheight{0.1}%
    \setlength\tabcolsep{0pt}%
    \centering
    \put(0, -0.113){\includegraphics[width=\unitlength,page=1]{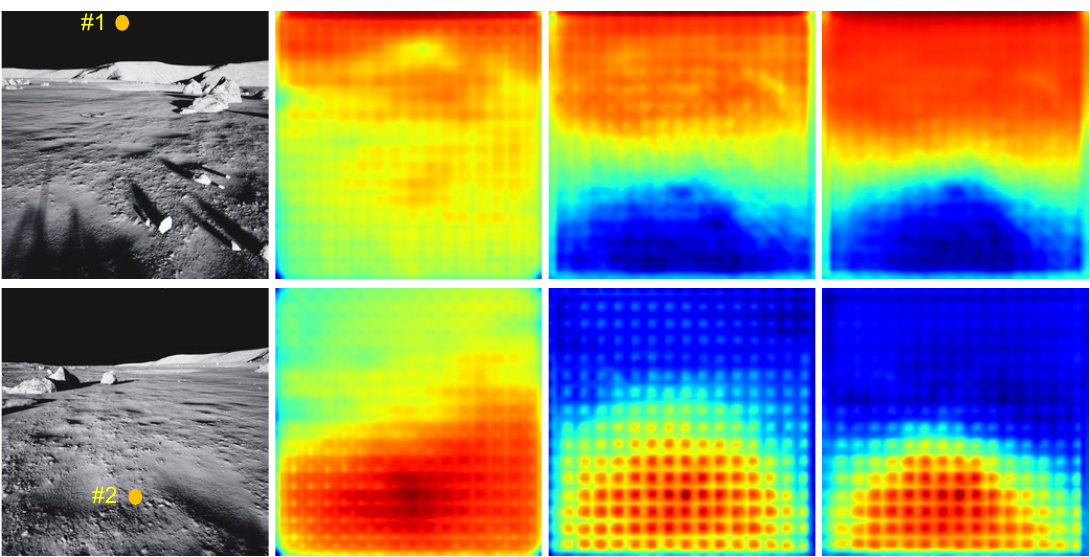}}%
    \put(0.12,0.4){\makebox(0,0)[t]{\lineheight{0.1}\smash{\footnotesize\begin{tabular}[t]{c}Ref.\end{tabular}}}}%
    % \put(0.22,0.90320912){\makebox(0,0)[t]{\lineheight{0.1}\smash{\begin{tabular}[t]{c}Target View\end{tabular}}}}%
    \put(0.37,0.4){\makebox(0,0)[t]{\lineheight{0.1}\smash{\footnotesize\begin{tabular}[t]{c} w/o fusion
\end{tabular}}}}%
    \put(0.63,0.4){\makebox(0,0)[t]{\lineheight{0.1}\smash{\footnotesize\begin{tabular}[t]{c}w/o U-Net\end{tabular}}}}%\textbf{}
    \put(0.87,0.4){\makebox(0,0)[t]{\lineheight{0.1}\smash{\footnotesize\begin{tabular}[t]{c}base\end{tabular}}}}%

    % \put(0.90890912,0.91320912){\makebox(0,0)[t]{\lineheight{1.25}\smash{\begin{tabular}[t]{c}pixelNeRF~\cite{pixelnerf}\end{tabular}}}}%
  \end{picture}%
\endgroup%

    \textbf{\vspace{3.45em}}
    \caption{\textbf{Visualization of features under different ablation settings.} The features extracted by the visual foundation model and their corresponding similarity maps in different ablation settings: w/o fusion, w/o U-Net, and base. Two points are selected (\#1 in the sky region and \#2 on the lunar regolith), and global cosine similarity is computed to assess feature compactness. The results demonstrate that, after full module processing, the features become more coherent and compact.}
    % \vspace{-1pt}
    \label{fig:unet}
\end{figure}

\noindent \textbf{Importance of multi-level feature fusion and U-Net.} As shown in Tab.~\ref{tab:ablations}, incorporating 2D U-Net consistently yielded notable benefits across different datasets: it provided a PSNR improvement of about 0.5 dB on the LuSNAR dataset and of nearly 1 dB on the MoonBlender dataset. This finding indicates that refining features with U-Net effectively enhances rendering quality. To visualize this effectiveness, we utilized cosine similarity to examine the relationships between points and pixels. As illustrated in Fig.\ref{fig:unet}, we selected one pixel point each in the sky and ground semantic regions to assess similarity across the entire feature map. In the LuSNAR dataset, before any feature fusion, all pixels exhibited highly similar features. After applying multi-level feature fusion, features belonging to the same category showed greater similarity and clustered in a tighter region. Further refining the features with U-Net made the distribution smoother and concentrated similar features into smaller areas. These observations suggest that multi-level feature fusion produces tighter representations, serving as advantageous priors for the subsequent cost volume construction and feature matching.

\begin{table}[t]
\centering
\small
\caption{\textbf{Ablation Studies.} This table summarizes the results of removing semantic fusion and the 2D U-Net from our pipeline. Comparing the performance with and without these components highlights their essential role in producing high-quality novel view synthesis.
% Both our epipolar encoder and our probabilistic sampling scheme are essential for high-quality novel view synthesis. Depth regularization slightly impacts rendering quality.
}
\begin{tabular}{l|ccc}
\toprule
Methods & PSNR $\uparrow$   & SSIM $\uparrow$   & LPIPS $\downarrow$    \\
\midrule
% \textcolor{black!50}{\textit{LuSNAR}}     &                   &                   &                       \\
\multicolumn{4}{l}{\textcolor{black!50}{\textit{LuSNAR}~\cite{liu2024lusnar}}} \\
MoonGS (Full)       & \textbf{23.80}    & \textbf{0.816}    &     \textbf{0.134}    \\
w/o U-Net            &  23.29            & 0.786             &              0.137    \\
w/o Semantic Fusion  &  23.02            & 0.781             &              0.141    \\
\midrule
% \textcolor{black!50}{\textit{MoonBlender}}&                   &                   &                       \\
\multicolumn{4}{l}{\textcolor{black!50}{\textit{MoonBlender}}} \\
MoonGS (Full)       & \textbf{26.08}    & \textbf{0.721}    &     \textbf{0.139}    \\
w/o U-Net            &  25.14            & 0.687             &              0.162    \\
w/o Semantic Fusion  &  25.90            & 0.719             &              0.140    \\
% No Probabilistic Sampling   &             24.62 &             0.828 &              0.171 \\
% Plus Depth Regularization    & \underline{25.94} & \underline{0.862} &  \underline{0.137} \\
\bottomrule
\end{tabular}
\vspace{-5pt}
\label{tab:ablations}
\end{table}

\begin{figure}[t]
    % \vspace{-1em}
    \centering
    \includegraphics[width=\linewidth]{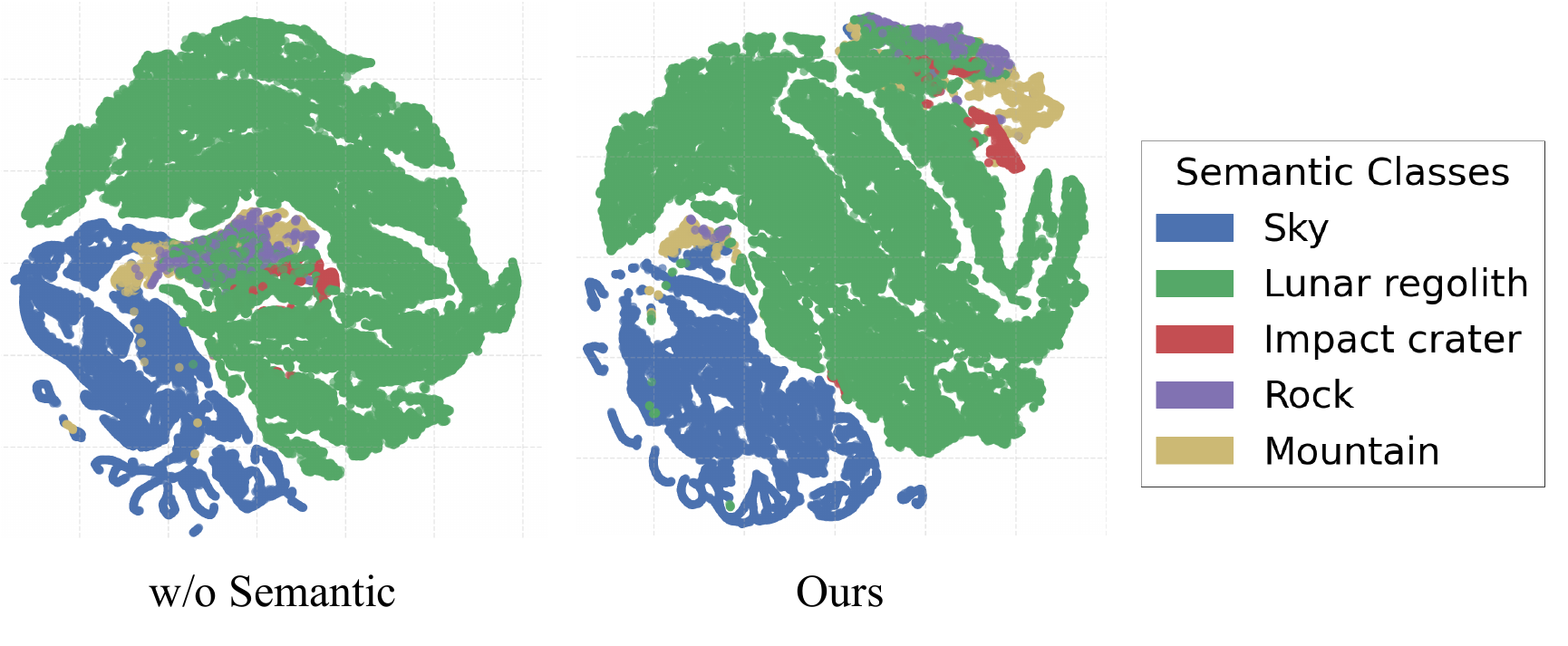}
    \caption{\textbf{t-SNE analysis of semantic fusion.} The t-SNE analysis highlights the impact of semantic fusion on feature distribution. The left plot (w/o semantic) shows the feature distribution without semantic fusion, while the right plot (ours) presents the features after semantic fusion. Each color corresponds to a different semantic class. Incorporating semantic information leads to a more distinct separation between features.}
    \label{fig:tsne}
\end{figure}

\noindent \textbf{Importance of semantic fusion.} As shown in Tab.~\ref{tab:ablations}, introducing semantic information provided certain enhancement in rendering quality: approximately 0.2 dB improvement on the MoonBlender dataset and about 0.8 dB on the LuSNAR dataset. We believe that the smaller improvement on MoonBlender is primarily due to the lower number of semantic categories. Nevertheless, incorporating semantic features still improved the model’s performance on various quality metrics. On the one hand, adding semantic information increased the network’s learnable parameters, enabling a more thorough mapping from pixels to Gaussian distributions. On the other hand, semantics encouraged different categories to form tighter clusters in feature space, rather than overlapping. Given the strong correlation between semantics and texture, we visualized the predicted Gaussian color parameters in the LuSNAR dataset. Fig.\ref{fig:tsne} indicates that, without semantic information, categories with similar colors (e.g., craters, lunar regolith, and rocks) became tangled in feature space. Adding semantic features, these categories' features became separated, effectively mitigating the entanglement. When semantic masks were unreliable, the fusion branch was gated off. Under this setting, the model still surpassed the baselines on both datasets in PSNR/SSIM/LPIPS, as shown in Tab.~\ref{tab:main_result} and Tab.~\ref{tab:ablations}. This indicated that, when semantics were unavailable, our model remained applicable to lunar scenes.

\noindent \textbf{Importance of ranking loss.} We observed that in the pixelSplat model, weakly textured regions tended to be represented as Gaussians with uncertain depth, whereas MVSplat failed to render such regions. This is mainly because weakly textured regions cannot be effectively handled by photometric loss alone. Two primary factors contribute to this issue. First, the strong fitting capability of Gaussians can lead the training to converge to local minima, such as MVSplat. Even if the Gaussian's center position is slightly inaccurate, its anisotropy allows for a reasonably good fit, rendering photometric loss insufficient for proper training. Second, in weakly textured areas, visual geometric priors are less pronounced, causing epipolar attention and MVS-based methods to easily fail. To address these challenges, we introduced a semantic-based ranking loss to regularize weakly textured regions, thereby achieving more meticulous and smoother depth estimates. We primarily validated this loss on the MoonBlender dataset, as it features notable weakly textured scenes compared to LuSNAR. As shown in Fig.\ref{fig:depth_compare}, pixelSplat yielded incorrect depth in the sky region, and MVSplat fell into local minima, producing erroneous depth estimates. Our method was able to produce good depth. Fig.\ref{fig:depthloss} shows that although our model could produce smooth depth on the ground before using ranking loss, there were still abnormal areas in the texture of the background sky. However, after regularization with ranking loss, this abnormality was resolved. Based on the foundation model and regularization loss, our method achieved better performance, which proves the effectiveness of our proposed method.

Further, we evaluated the extensibility of this approach. We integrated the ranking loss into the pixelSplat method to regularize. As shown in Fig.\ref{fig:depthloss}, the sky depth became noticeably more reliable after regularization, demonstrating that our method can be effectively incorporated into related approaches. This finding further underscores the extensibility and effectiveness of our proposed ranking loss function. When advanced models are used to reconstruct lunar scenes, our proposed semantic ranking loss is expected to extend to these models, thereby improving their effect.

\begin{table}[t]
\centering
\small
\vspace{-1em}
\caption{\textbf{Ablation study of the heuristic resampling strategy.} 
This table shows the impact of our heuristic resampling strategy on the performance of various models, evaluated after synthesizing novel views for each model. Superscript values denote the specific resampling position. Overall, our strategy enhances the final rendering quality.}
\begin{tabular}{l|ccc}
\toprule
Methods & PSNR $\uparrow$ & SSIM $\uparrow$ & LPIPS $\downarrow$ \\
\midrule
\multicolumn{4}{l}{\textcolor{black!50}{\textit{MoonGS}}}  \\
% \midrule
w/o Resample                        & 22.46 & 0.767 & \textbf{0.151} \\
Fixed Resample\textsuperscript{1}   & 22.76 & 0.795 & 0.186 \\
Fixed Resample\textsuperscript{4}   & \underline{23.97} & 0.848 & 0.162 \\
Fixed Resample\textsuperscript{7}   & 23.88 & \underline{0.850} & 0.157 \\
Fixed Resample\textsuperscript{11}  & 23.38 & 0.812 & 0.175 \\
w/ Resample       & \textbf{24.09} & \textbf{0.854} & \underline{0.155} \\
\midrule
\multicolumn{4}{l}{\textcolor{black!50}{\textit{pixelSplat}~\cite{charatan2024pixelsplat}}} \\
% \midrule
w/o Resample      & 17.92 & 0.481 & 0.383 \\
w/ Resample       & \textbf{19.20} & \textbf{0.556} & \textbf{0.349} \\
\bottomrule
\end{tabular}
% \vspace{-5pt}
\label{tab:resample}
\end{table}

\noindent \textbf{Importance of heuristic resampling strategy.} We evaluated the resampling positions under a fixed-interval resampling approach. Tab.\ref{tab:resample} indicated that, our resampling strategy outperformed the fixed-interval method on the LuSNAR video sequence dataset, achieving the highest PSNR of 24.09 dB. The principal advantages of our strategy include low reliance on prior information and high computational efficiency. Notably, the high computational efficiency stems evaluation of resampling positions being performed in parallel with the rendering process.

We also applied the resampled positions to pixelSplat. Tab.~\ref{tab:resample} shows that our method improved the performance of other models. This nicely illustrates that our method is both effective and scalable.

\section{Discussions}

This section presents the scope of MoonGS, states its limitations, and outlines future work.

First, several modules in our pipeline use semantics. When semantics are unavailable, we activate a fallback mechanism that preserves stability. In future work, we plan to reduce this dependence by incorporating priors that are easier to obtain. These include edge cues, regionness cues, uncertainty-aware pseudo-labels, and surface normals. 

Second, MoonGS focuses on pair-based feed-forward novel view synthesis. We specifically adopt this setting to address the challenge of sparse viewpoints in lunar exploration. However, the framework can be extended to incorporate three or more views, as the underlying vision foundation model supports multi-image inputs and the cost volume formulation generalizes to pairwise matching. In each forward pass the number of Gaussian primitives equals the number of input pixels, so the per-pass memory footprint depends on image resolution and per-primitive state rather than terrain size. Larger maps can be obtained by fusing outputs from multiple inferences of our model. For the fused maps, exploring hierarchical tiled representations and spatially adaptive splatting can improve memory efficiency and scalability, which we leave to future work. 

Third, although inference is fast, the network is not yet lightweight enough for constrained devices. We will explore model distillation, pruning, and quantization to obtain smaller models while preserving accuracy. 

Fourth, we keep the default backbone (DUSt3R) fixed during training. With limited lunar data, this choice improves stability and data efficiency. As larger and more diverse datasets become available, we plan to partially unfreeze the backbone and fine-tune selected layers to further improve adaptation. 

Fifth, as validated by the VGGT integration, our framework exhibits robust applicability to advanced backbone architectures. By incorporating stronger priors while freezing the foundation model, we ensure parameter efficiency without sacrificing performance. Future work will leverage this adaptable design to integrate next-generation foundation models, a strategy that guarantees our method continues to capitalize on the rapid advancements in underlying feature representations.

Finally, the current pipeline assumes known relative camera poses as input. In future work, we will investigate pose-free variants to broaden the applicability of our method.

\section{Conclusion}
We introduced MoonGS, a feed-forward 3DGS method tailored to lunar environments. It aimed to improve the perception ability of robots on lunar terrains by addressing the problems of invalid epipolar constraints, ambiguous depth in weakly textured regions and insufficient data observations. First, we employed robust depth features provided by the vision foundation model to achieve better Gaussian positions, particularly in regions with limited overlapping views. Second, we incorporated lunar semantic priors into the model, which significantly enhanced its overall performance. To address ambiguous depth in regions with weak texture, we proposed a semantic-based ranking loss for depth regularization. In addition, we designed a heuristic resampling strategy to further improve the rendering quality. Finally, to facilitate a comprehensive evaluation under challenging conditions, we introduce MoonBlender, a dataset extending existing lunar datasets and specifically designed to assess reconstruction performance in weakly textured regions. We also reported qualitative results on real Chang’e imagery, where MoonGS delivered higher-fidelity reconstructions and showed robustness under varying illumination. Experimental results showed that compared to state-of-the-art methods~\cite{charatan2024pixelsplat, chen2024mvsplat}, MoonGS achieved higher-quality reconstructions on two lunar datasets, even in non-overlapping or weakly textured regions. 
% \revision{Furthermore, experiments with the advanced VGGT~\cite{wang2025vggtvisualgeometrygrounded} backbone demonstrated the scalability of our framework.} 
Furthermore, experiments with the advanced VGGT~\cite{wang2025vggtvisualgeometrygrounded} backbone validated the broad applicability of our framework, demonstrating its capacity to seamlessly leverage state-of-the-art architectures for enhanced performance.
The synthesized novel viewpoints have improved the perception capability of the lunar rover, thereby enabling more efficient and safer lunar exploration missions.

\bibliographystyle{IEEEtran}
\bibliography{references}

% \newpage

 

% \end{IEEEbiography}



\vfill

\end{document}